\pdfoutput=1
\documentclass{article} %
\usepackage{iclr2027_conference,times}

\PassOptionsToPackage{hyphens}{url}
\usepackage{fontawesome5}
\usepackage{hyperref}
\usepackage{url}
\definecolor{darkblue}{rgb}{0, 0, 0.5}
\hypersetup{colorlinks=true, citecolor=darkblue, linkcolor=darkblue, urlcolor=darkblue}

\usepackage[T1]{fontenc}
\usepackage[utf8]{inputenc}
\usepackage{microtype}
\usepackage{inconsolata}
\usepackage{graphicx}
\usepackage{wrapfig}
\usepackage{amsmath}
\usepackage{amssymb}
\usepackage{booktabs}
\usepackage{tabularx}
\usepackage{multirow}
\usepackage{makecell}
\usepackage{xcolor}
\usepackage{colortbl}
\usepackage[most]{tcolorbox}
\usepackage{pgfplots}
\pgfplotsset{compat=1.18}
\usetikzlibrary{calc,arrows.meta}

\usepackage{etoolbox}
\AtBeginEnvironment{table}{\setlength{\abovecaptionskip}{0pt}\setlength{\belowcaptionskip}{6pt}}
\AtBeginEnvironment{table*}{\setlength{\abovecaptionskip}{0pt}\setlength{\belowcaptionskip}{6pt}}
\AtBeginEnvironment{wraptable}{\setlength{\abovecaptionskip}{0pt}\setlength{\belowcaptionskip}{6pt}}

\newcommand{\qatTableSetup}{%
  \setlength{\tabcolsep}{2pt}%
  \setlength{\heavyrulewidth}{0.9pt}%
  \setlength{\lightrulewidth}{0.35pt}%
  \setlength{\cmidrulewidth}{0.3pt}%
  \setlength{\aboverulesep}{1.4pt}%
  \setlength{\belowrulesep}{1.4pt}}
\newcommand{\qatTableHead}[1]{\textbf{#1}}
\newcommand{\qatTableModel}[1]{\textbf{\textit{#1}}}
\newcommand{\qatTableShade}{\rowcolor{black!7}}
\newcolumntype{Q}[1]{>{\hsize=#1\hsize\centering\arraybackslash}X}
\newcolumntype{Y}{@{}l<{\hspace{4pt}}c<{\hspace{2pt}}Q{1.0}Q{.95}Q{.9}Q{.95}Q{.98}Q{1.2}Q{.98}Q{1.14}Q{.95}Q{.95}@{}}
\newcommand{\qatTableHeaderRule}{\specialrule{0.35pt}{1.5pt}{1pt}\specialrule{0.35pt}{0pt}{2pt}}
\newcommand{\qatTableModelRule}{\midrule[0.6pt]}

\newcolumntype{Y}{@{}l<{\hspace{4pt}}c<{\hspace{2pt}}Q{1.052}Q{.989}Q{.870}Q{.926}Q{1.008}Q{1.198}Q{1.068}Q{1.147}Q{.870}Q{.872}@{}}
\newcommand{\scorecap}[2]{#1\,(#2)}
\newcommand{\bestscorecap}[2]{\textbf{#1}\,(#2)}
\newcommand{\ours}{OnQ}
\newcommand{\blackcircled}[1]{%
  \tikz[baseline=(num.base)]{
    \node[
      circle,
      fill=black,
      text=white,
      inner sep=0pt,
      minimum size=1em,
      font=\sffamily\bfseries
    ] (num) {#1};
  }%
}

\title{Right On Q: Recovering Reasoning under\\Quantization via Self-Generated Trajectories}

\author{Boyuan Zheng$^{1}$, Jing Xiong$^{2}$, Chenwei Wu$^{1}$, Hui Shen$^{1}$, \\
\textbf{Zesen Zhao$^{1}$, Hei Ting Una Chan$^{1}$, Xueshen Liu$^{1}$, Zhen Zhang$^{2}$,} \\
\textbf{Liyue Shen$^{1,\dagger}$, Zechun Liu$^{3,\dagger,\ddagger}$, Z. Morley Mao$^{1,\dagger,\ddagger}$} \\
$^{1}$University of Michigan, Ann Arbor \quad $^{2}$University of California, Santa Barbara \quad $^{3}$Meta AI}

\iclrfinalcopy
\begin{document}

\maketitle
{\renewcommand{\thefootnote}{\fnsymbol{footnote}}\footnotetext[2]{Corresponding authors.}\footnotetext[3]{Equal advising.}}
\lhead{Preprint}

\begin{abstract}
Quantization-aware training (QAT) aims to recover reasoning capabilities in low-bit large language models (LLMs). Existing recipes typically supervise the quantized student with the original full-precision model on fixed response sequences. However, the resulting low-bit quantized models still struggle on long-horizon reasoning tasks and often fail to terminate within a token budget that suffices for the full-precision model. We identify two challenges in closing this gap. First, fixed-response supervision leaves a mismatch between training prefixes and those encountered during student generation, and this \textbf{trajectory divergence} grows as responses lengthen. Second, on its own prefixes, the student underpredicts some next-token candidates and overpredicts others, and this \textbf{distributional misalignment} becomes more frequent at high-entropy prefixes. We present \textbf{\ours{}}, a low-bit recovery pipeline that combines on-policy distillation and candidate-wise Dynamic KL. Starting from block-wise calibration, the quantized student generates responses and receives full-precision teacher supervision on its own prefixes. Dynamic KL weights forward and reverse KL by each candidate's teacher--student probability ratio to balance correction of under- and overprediction. Across Qwen3-1.7B, 4B, and 8B at 2- and 3-bit precision, \ours{} achieves the highest nine-benchmark average, with the largest gains on long-horizon reasoning benchmarks. At 2 bits, it outperforms every baseline on every benchmark at every model size, improving the Qwen3-8B average by 10.26 points over the strongest baseline. It also produces shorter responses with fewer truncations and repetitive errors. The code is available \href{https://github.com/Boyuannz/onq}{\mbox{\faGithub\,here}}.
\end{abstract}
\section{Introduction}
\begin{figure*}[!t]
\centering
\resizebox{\textwidth}{!}{\begingroup%
\endlinechar=-1\relax\input{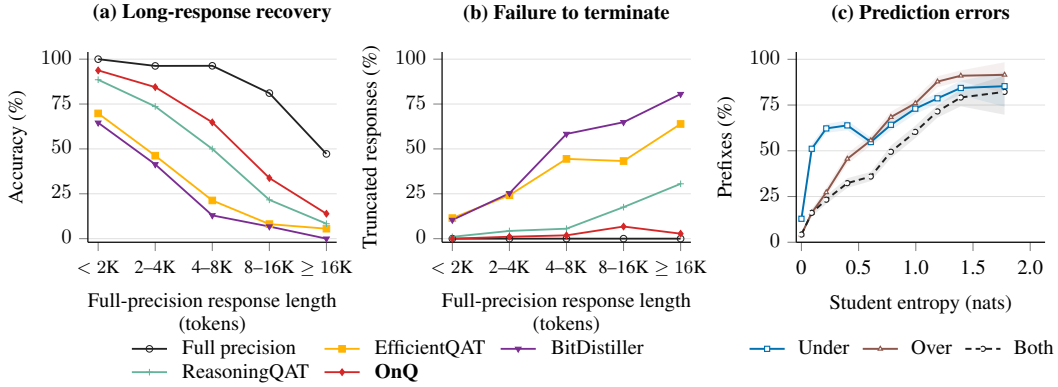}\endlinechar=13\relax%
\definecolor{figEfficient}{HTML}{FFB300}%
\definecolor{figBitDistiller}{HTML}{7B2C93}%
\definecolor{figReasoning}{HTML}{65B39A}%
\definecolor{figOnQ}{HTML}{D62728}%
\definecolor{figOver}{HTML}{8C564B}%
\begin{tikzpicture}[/pgfplots/every axis/.append style={
 scale only axis,width=4.25cm,height=3.25cm,
 tick label style={font=\fontsize{10}{11}\selectfont},
 label style={font=\fontsize{10}{11}\selectfont},
 title style={font=\fontsize{10}{11}\selectfont\bfseries,yshift=5pt},
 axis x line*=bottom,axis y line*=left,axis line style={line width=.6pt},
 tick style={line width=.5pt},tick align=outside,
 ymajorgrids,grid style={black!12,line width=.35pt},
 scaled ticks=false,typeset ticklabels with strut,
 legend style={font=\fontsize{10}{11}\selectfont,draw=none,fill=none,
  cells={anchor=west},row sep=-1pt,column sep=3pt},
 clip=false}]
\begin{axis}[name=fa,xmin=-.18,xmax=4.18,ymin=-2,ymax=104,
 ytick={0,25,50,75,100},xtick={0,1,2,3,4},
 xticklabels={$<2$K,2--4K,4--8K,8--16K,$\geq16$K},
 x tick label style={font=\fontsize{9}{10}\selectfont},
 ylabel={Accuracy (\%)},xlabel={Full-precision response length\\(tokens)},
 xlabel style={align=center},title={(a) Long-response recovery},
 legend style={at={(1.2118,-.47)},anchor=north,legend columns=3,column sep=5pt}]
\addplot[qatRef,line width=.9pt,mark=o,mark size=1.6pt,mark options={draw=qatRef,fill=white,line width=.5pt}] coordinates {(0,100.000000000000) (1,96.236559139785) (2,96.296296296296) (3,81.081081081081) (4,47.222222222222)};
\addlegendentry{Full precision}
\addplot[figEfficient,line width=.9pt,mark=square*,mark size=1.6pt,mark options={draw=figEfficient,fill=figEfficient,line width=.5pt}] coordinates {(0,69.791666666667) (1,46.236559139785) (2,21.296296296296) (3,8.108108108108) (4,5.555555555556)};
\addlegendentry{EfficientQAT}
\addplot[figBitDistiller,line width=.9pt,mark=triangle*,mark size=1.6pt,mark options={draw=figBitDistiller,fill=figBitDistiller,line width=.5pt,rotate=180}] coordinates {(0,64.583333333333) (1,41.397849462366) (2,12.962962962963) (3,6.756756756757) (4,0.000000000000)};
\addlegendentry{BitDistiller}
\addplot[figReasoning,line width=.9pt,mark=+,mark size=1.6pt,mark options={draw=figReasoning,fill=figReasoning,line width=.5pt}] coordinates {(0,88.541666666667) (1,73.655913978495) (2,50.000000000000) (3,21.621621621622) (4,8.333333333333)};
\addlegendentry{ReasoningQAT}
\addplot[figOnQ,line width=.9pt,mark=diamond*,mark size=1.6pt,mark options={draw=figOnQ,fill=figOnQ,line width=.5pt}] coordinates {(0,93.750000000000) (1,84.408602150538) (2,64.814814814815) (3,33.783783783784) (4,13.888888888889)};
\addlegendentry{\textbf{OnQ}}
\end{axis}
\begin{axis}[name=fb,at={(fa.east)},anchor=west,xshift=1.8cm,
 xmin=-.18,xmax=4.18,ymin=-2,ymax=104,
 ytick={0,25,50,75,100},xtick={0,1,2,3,4},
 xticklabels={$<2$K,2--4K,4--8K,8--16K,$\geq16$K},
 x tick label style={font=\fontsize{9}{10}\selectfont},
 ylabel={Truncated responses (\%)},xlabel={Full-precision response length\\(tokens)},
 xlabel style={align=center},title={(b) Failure to terminate}]
\addplot[qatRef,line width=.9pt,mark=o,mark size=1.6pt,mark options={draw=qatRef,fill=white,line width=.5pt}] coordinates {(0,0.000000000000) (1,0.000000000000) (2,0.000000000000) (3,0.000000000000) (4,0.000000000000)};
\addplot[figEfficient,line width=.9pt,mark=square*,mark size=1.6pt,mark options={draw=figEfficient,fill=figEfficient,line width=.5pt}] coordinates {(0,11.458333333333) (1,24.193548387097) (2,44.444444444444) (3,43.243243243243) (4,63.888888888889)};
\addplot[figBitDistiller,line width=.9pt,mark=triangle*,mark size=1.6pt,mark options={draw=figBitDistiller,fill=figBitDistiller,line width=.5pt,rotate=180}] coordinates {(0,10.416666666667) (1,25.268817204301) (2,58.333333333333) (3,64.864864864865) (4,80.555555555556)};
\addplot[figReasoning,line width=.9pt,mark=+,mark size=1.6pt,mark options={draw=figReasoning,fill=figReasoning,line width=.5pt}] coordinates {(0,1.041666666667) (1,4.301075268817) (2,5.555555555556) (3,17.567567567568) (4,30.555555555556)};
\addplot[figOnQ,line width=.9pt,mark=diamond*,mark size=1.6pt,mark options={draw=figOnQ,fill=figOnQ,line width=.5pt}] coordinates {(0,0.000000000000) (1,1.075268817204) (2,1.851851851852) (3,6.756756756757) (4,2.777777777778)};
\end{axis}
\begin{axis}[name=fc,at={(fb.east)},anchor=west,xshift=1.8cm,
 xmin=-.045,xmax=2.13,ymin=0,ymax=104,ytick={0,25,50,75,100},
 xtick={0,.5,1,1.5,2},xticklabels={0,0.5,1.0,1.5,2.0},
 ylabel={Prefixes (\%)},xlabel={Student entropy (nats)},
 title={(c) Prediction errors},
 legend style={at={(.43,-.47)},anchor=north,legend columns=3,column sep=4pt}]
\addplot[draw=none,fill=qatBlue,fill opacity=.09,forget plot] coordinates {(0.000299329823,13.889259641801) (0.090484856869,53.772356758468) (0.217373497148,64.684661592274) (0.401429080905,66.614438231917) (0.606526626474,57.263871488627) (0.782314363116,67.149060573352) (0.995462807241,75.845002364825) (1.189575251928,81.684981684982) (1.393218959106,88.304093567251) (1.775990895392,93.195876288660) (1.775990895392,74.049205476537) (1.393218959106,80.446478050534) (1.189575251928,75.595049728752) (0.995462807241,70.114862261607) (0.782314363116,61.175235485115) (0.606526626474,52.324668058552) (0.401429080905,61.190494486971) (0.217373497148,59.780199579832) (0.090484856869,48.498052879688) (0.000299329823,11.841156562197)} \closedcycle;
\addplot[draw=none,fill=figOver,fill opacity=.09,forget plot] coordinates {(0.000299329823,4.737565492247) (0.090484856869,18.129283439264) (0.217373497148,29.909559076307) (0.401429080905,48.512419327720) (0.606526626474,58.430074393730) (0.782314363116,71.393647985164) (0.995462807241,79.019252160957) (1.189575251928,90.702426787150) (1.393218959106,93.956085455330) (1.775990895392,98.397101020222) (1.775990895392,81.202757302855) (1.393218959106,87.958041855486) (1.189575251928,84.712174899799) (0.995462807241,72.702312559298) (0.782314363116,65.499510589739) (0.606526626474,52.891445403383) (0.401429080905,42.827552644010) (0.217373497148,24.770526614752) (0.090484856869,14.152212823673) (0.000299329823,3.679193327900)} \closedcycle;
\addplot[draw=none,fill=qatRef,fill opacity=.09,forget plot] coordinates {(0.000299329823,4.746207821932) (0.090484856869,18.094601651490) (0.217373497148,25.672753062540) (0.401429080905,35.048489636520) (0.606526626474,38.467166512770) (0.782314363116,52.887048275957) (0.995462807241,63.719176663307) (1.189575251928,74.920891859022) (1.393218959106,83.679550980877) (1.775990895392,91.011300678300) (1.775990895392,69.676896084708) (1.393218959106,74.277245377413) (1.189575251928,68.141592920354) (0.995462807241,56.885371025118) (0.782314363116,45.998315080034) (0.606526626474,33.424406682669) (0.401429080905,29.656634255072) (0.217373497148,20.928269684682) (0.090484856869,14.073461628445) (0.000299329823,3.696095890338)} \closedcycle;
\addplot[qatBlue,solid,line width=.95pt,mark=square*,mark size=1.25pt,mark options={fill=white,draw=qatBlue,line width=.55pt}] coordinates {(0.000299329823,12.841456882700) (0.090484856869,51.096956829441) (0.217373497148,62.222222222222) (0.401429080905,63.880597014925) (0.606526626474,54.784366576819) (0.782314363116,64.195450716091) (0.995462807241,73.015873015873) (1.189575251928,78.673835125448) (1.393218959106,84.347826086957) (1.775990895392,85.300207039337)};
\addlegendentry{Under}
\addplot[figOver,solid,line width=.95pt,mark=triangle*,mark size=1.25pt,mark options={fill=white,draw=figOver,line width=.55pt}] coordinates {(0.000299329823,4.204606320300) (0.090484856869,16.065109695683) (0.217373497148,27.259259259259) (0.401429080905,45.597014925373) (0.606526626474,55.660377358491) (0.782314363116,68.407750631845) (0.995462807241,75.925925925926) (1.189575251928,87.813620071685) (1.393218959106,91.014492753623) (1.775990895392,91.511387163561)};
\addlegendentry{Over}
\addplot[qatRef,densely dashed,line width=.95pt,mark=*,mark size=1.25pt,mark options={fill=white,draw=qatRef,line width=.55pt}] coordinates {(0.000299329823,4.204606320300) (0.090484856869,16.065109695683) (0.217373497148,23.259259259259) (0.401429080905,32.313432835821) (0.606526626474,35.916442048518) (0.782314363116,49.452401010952) (0.995462807241,60.317460317460) (1.189575251928,71.505376344086) (1.393218959106,79.130434782609) (1.775990895392,82.194616977226)};
\addlegendentry{Both}
\end{axis}
\pgfresetboundingbox
\path[use as bounding box] ([xshift=-1.45cm,yshift=-2.35cm]fa.south west) rectangle ([xshift=.15cm,yshift=.9cm]fc.north east);
\end{tikzpicture}
\endgroup}
\caption{\textbf{Low-bit fixed-response QAT fails on long-horizon reasoning; \ours{} outperforms all QAT baselines (Qwen3-1.7B W2).} (a) MATH-500 accuracy and (b) share of responses that reach the token limit, grouped by full-precision response length. (c) Share of prefixes with underpredicted candidates, overpredicted candidates, or both, by student entropy.}
\label{fig:opd-motivation}
\end{figure*}

\begingroup
Low-bit quantization reduces the memory footprint of large language models (LLMs), making their deployment more practical~\citep{frantar2022gptq,DBLP:conf/acl/ChenSXWGZ025}. Preserving reasoning capabilities at low precision remains challenging: prior work reports substantial accuracy loss~\citep{liu2025quantization} and inefficient generation~\citep{DBLP:journals/corr/abs-2606-00206} in quantized reasoning models. Quantization-aware training (QAT) addresses this loss by adapting model parameters through quantized forward passes, often with supervision from the original full-precision model~\citep{DBLP:journals/corr/HintonVD15,DBLP:conf/acl/LiuO0CSMSKC24,DBLP:conf/acl/DuZCGCCX24,DBLP:conf/iclr/OkoshiOFM26}. These distillation recipes typically train on fixed response sequences. Their effectiveness therefore depends on how well the training prefixes cover those encountered by the quantized student during inference~\citep{DBLP:conf/iclr/AgarwalVZSGGB24}.

To examine the remaining challenges in low-bit reasoning, we analyze Qwen3-1.7B at 2-bit precision (\hyperref[fig:opd-motivation]{Figure~\ref*{fig:opd-motivation}}). On MATH-500, QAT baselines recover little of the full-precision model's accuracy on questions eliciting long full-precision responses (\hyperref[fig:opd-motivation]{Figure~\ref*{fig:opd-motivation}.a}). They also frequently reach the token limit on these questions, while the full-precision model terminates within the same budget (\hyperref[fig:opd-motivation]{Figure~\ref*{fig:opd-motivation}.b}). Fixed-response distillation supervises only reference prefixes, but quantization changes how the student generates: the longer the response, the more the student conditions on its own outputs and the further its context drifts from those prefixes. This trajectory divergence motivates teacher supervision on the student's own trajectories.

Training on the student's own prefixes raises a second challenge: distributional misalignment. On these prefixes, the calibrated student exhibits two-sided probability errors: it substantially underpredicts some candidates and overpredicts others at the same prefix. These two-sided errors become more frequent at high-entropy prefixes (\hyperref[fig:opd-motivation]{Figure~\ref*{fig:opd-motivation}.c}). Prior work also finds that quantized models oversample overthinking markers at high-entropy positions~\citep{DBLP:journals/corr/abs-2606-00206}. Neither Kullback--Leibler (KL) direction alone corrects both sides equally well: forward KL divergence more reliably moves severely underpredicted candidates toward the teacher, while reverse KL does so for severely overpredicted candidates (\hyperref[sec:distillation-objective]{Section~\ref*{sec:distillation-objective}}). This calls for an objective that balances correction across both sides of the distribution.

Motivated by these findings, we introduce \ours{}, a low-bit recovery pipeline that combines on-policy distillation (OPD) and candidate-wise Dynamic KL~\citep{DBLP:conf/emnlp/JungYKL25}. Starting from a block-wise calibrated checkpoint, the quantized student generates responses and receives supervision from a frozen full-precision teacher on those same prefixes~\citep{DBLP:conf/iclr/AgarwalVZSGGB24}. Dynamic KL weights forward and reverse KL according to each candidate's teacher--student probability ratio, balancing the correction of under- and overpredicted probabilities along these evolving trajectories. On MATH-500, \ours{} recovers more accuracy than every QAT baseline at every response length and rarely reaches the token limit (\hyperref[fig:opd-motivation]{Figure~\ref*{fig:opd-motivation}.a--b}).

The key contributions of this study are as follows: \blackcircled{1} We identify two challenges, trajectory divergence and distributional misalignment, that limit low-bit reasoning recovery. \blackcircled{2} We present \ours{}, a low-bit recovery pipeline combining on-policy distillation and candidate-wise Dynamic KL. \blackcircled{3} We show that \ours{} achieves the highest nine-benchmark average in all six settings spanning Qwen3-1.7B, 4B, and 8B at 2- and 3-bit precision, improving the Qwen3-8B average at 2 bits by 10.26 points over the strongest baseline. It also improves generation behavior, producing shorter responses with fewer repetitive errors, and outperforms fixed-response distillation.

\par
\endgroup

\section{Related Work}
\paragraph{Low-bit reasoning recovery.}
Aggressive post-training quantization (PTQ)~\citep{frantar2022gptq,DBLP:conf/emnlp/ChengZSCHLL24} can impair reasoning capabilities~\citep{liu2025quantization,DBLP:journals/corr/abs-2606-00206}. QAT recovers low-bit accuracy by adapting model parameters through quantized forward passes, often with supervision from a full-precision teacher~\citep{DBLP:conf/acl/LiuO0CSMSKC24,DBLP:conf/acl/DuZCGCCX24,liu2026paretoq,DBLP:journals/corr/abs-2601-14888}. EfficientQAT uses block-wise optimization followed by end-to-end refinement of quantization step sizes~\citep{DBLP:conf/acl/ChenSXWGZ025}, while ReasoningQAT~\citep{DBLP:conf/iclr/OkoshiOFM26} combines mixed-domain calibration with teacher-guided fine-tuning to preserve reasoning capabilities acquired after pre-training. LLM-QAT~\citep{DBLP:conf/acl/LiuO0CSMSKC24}, BitDistiller~\citep{DBLP:conf/acl/DuZCGCCX24}, and ReasoningQAT compute teacher supervision on fixed training sequences rather than on the student's own generations.

\paragraph{On-policy distillation.}
Training on fixed prefixes while generating from the model's own predictions at inference causes exposure bias, which imitation learning and sequence modeling mitigate by training on the learner's own outputs~\citep{DBLP:journals/jmlr/RossGB11,DBLP:conf/nips/BengioVJS15,DBLP:journals/corr/RanzatoCAZ15}. OPD makes the student's current behavior part of the training distribution: the teacher supervises prefixes generated by the student itself~\citep{DBLP:conf/emnlp/LinWCL20,DBLP:conf/iclr/AgarwalVZSGGB24,lu2025opd}. Alongside the source of training prefixes, the distillation objective determines how teacher--student discrepancies are weighted: BitDistiller mixes forward and reverse KL with a precomputed teacher-confidence coefficient~\citep{DBLP:conf/acl/DuZCGCCX24}, while adaptive objectives adjust this balance using distribution discrepancies, teacher entropy, or candidate-wise probability ratios~\citep{DBLP:conf/coling/WuTWY0W25,pmlr-v306-jin26e,DBLP:conf/emnlp/JungYKL25}. Related recipes combine distillation with reinforcement learning~\citep{DBLP:journals/corr/abs-2506-02208} or run reinforcement learning on quantized models~\citep{DBLP:conf/iclr/HuangGYXMLYLCYL26,DBLP:conf/acl/GuWLLZLXWYLHG26,DBLP:journals/corr/abs-2601-14888}. \ours{} applies candidate-wise Dynamic KL on quantized student rollouts, adapting distillation feedback to teacher--student probability discrepancies.

\section{Method}
\label{sec:method}
\begin{figure}[!t]
\centering
\includegraphics[width=\textwidth]{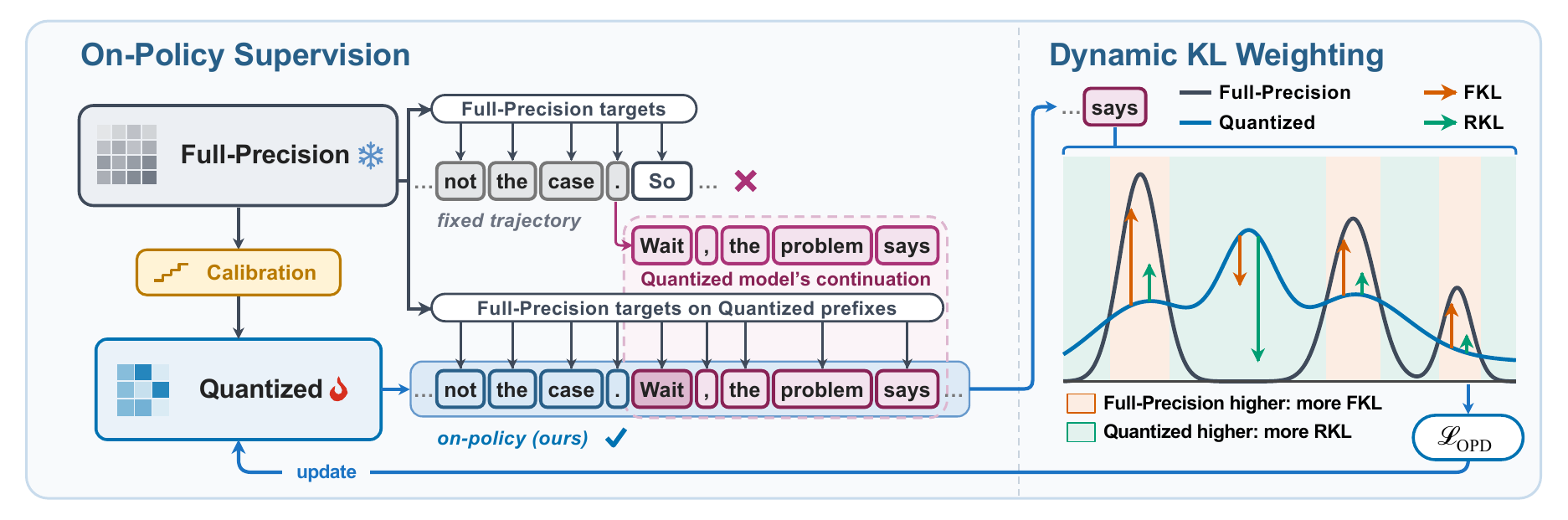}
\caption{Overview of \ours{}. Left: On-policy distillation trains the calibrated student on its own generated prefixes. Right: Dynamic KL weights forward and reverse KL by each candidate's teacher--student probability ratio.}
\label{fig:method-overview-placeholder}
\end{figure}

\ours{} targets low-bit capability recovery through teacher supervision along the quantized student's own reasoning trajectories (\hyperref[fig:method-overview-placeholder]{Figure~\ref*{fig:method-overview-placeholder}}). Block-wise calibration provides the initialization for on-policy training. The student then samples responses through its quantized forward pass, and its full-precision teacher supplies next-token distributions conditioned on the same prefixes; this on-policy supervision addresses trajectory divergence. To address distributional misalignment, we align the student with these distributions using candidate-wise Dynamic KL, which weights forward and reverse KL (FKL and RKL) according to the teacher--student probability ratio for each candidate token.

\subsection{Quantization and Initialization}
\label{sec:weight-quantization}
\label{sec:quantized-student}

We use uniform asymmetric weight quantization~\citep{DBLP:conf/acl/ChenSXWGZ025,DBLP:conf/iclr/OkoshiOFM26}. A full-precision latent weight $w$ is mapped to an $N$-bit code $w_{\mathrm{int}}$ and a dequantized weight $\hat w$:
\par
\begin{equation}
\begin{aligned}
w_{\mathrm{int}} &= \operatorname{clamp}(\lfloor w/s\rceil+z,0,2^N-1),\\
\hat w &= s(w_{\mathrm{int}}-z).
\end{aligned}
\label{eq:weight-quantizer}
\end{equation}
\noindent Here $s>0$ is the scale, $z$ is the integer zero point used in the forward pass, and $N$ is the target bit width; $\lfloor\cdot\rceil$ rounds to the nearest integer, and $\operatorname{clamp}(x,a,b)$ clips $x$ to $[a,b]$. The scale and zero point are shared within each group of weights. The quantized forward pass uses $\hat w$, while training updates the latent weight $w$.

We initialize the student with block-wise calibration, following Block-AP in EfficientQAT~\citep{DBLP:conf/acl/ChenSXWGZ025}. Starting from round-to-nearest (RTN) quantization with clipped per-group ranges, we train each transformer block $f_l$ in sequence on the calibration set $\mathcal C$, updating its weights $W_l$, scales $s_l$, and zero points $z_l$ with a block-wise reconstruction loss:
\begin{equation}
\min_{W_l,s_l,z_l}\;\mathbb E_{\mathcal C}\!\left[\operatorname{MSE}\!\left(f_l(\hat X_{l-1};\hat W_l),X_l^{\mathrm{FP}}\right)\right].
\label{eq:calibration-objective}
\end{equation}
Here $\hat W_l$ follows \hyperref[eq:weight-quantizer]{Equation~\ref*{eq:weight-quantizer}}, $\hat X_{l-1}$ is the input from preceding calibrated blocks, and $X_l^{\mathrm{FP}}$ is the output of the full-precision block on full-precision inputs. This calibrated checkpoint provides the starting policy for on-policy distillation (\hyperref[app:quantization-training]{Appendix~\ref*{app:quantization-training}}).

\subsection{On-Policy Distillation with Dynamic KL}
\label{sec:student-trajectories}
\label{sec:distillation-objective}

\paragraph{Student-generated prefixes.}
Let $Q_\theta$ be the quantized student with trainable $\theta=(W,s)$ and fixed zero points $z$, and $P$ the frozen full-precision teacher. For a prompt $x$, the student samples a response $y$ through its quantized forward pass. The teacher supplies next-token distributions at each generated prefix $h_t=(x,y_{<t})$, including prefixes containing earlier student errors~\citep{DBLP:conf/iclr/AgarwalVZSGGB24} (\hyperref[fig:method-overview-placeholder]{Figure~\ref*{fig:method-overview-placeholder}}, left).

\begin{figure}[t]
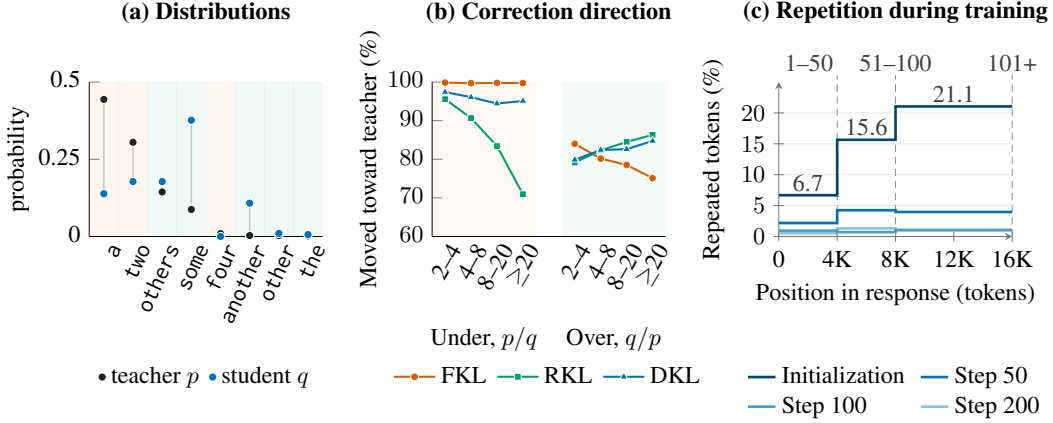

\centering
\resizebox{\textwidth}{!}{\begingroup%
\begin{tikzpicture}[baseline=(pa.south),every axis/.append style={scale only axis,height=2.25cm,tick label style={font=\fontsize{10}{11}\selectfont},label style={font=\fontsize{10}{11}\selectfont},title style={font=\fontsize{10}{11}\selectfont\bfseries,yshift=15pt},axis x line*=bottom,axis y line*=left,clip=false,
legend style={font=\fontsize{10}{11}\selectfont,draw=none,fill=none,legend columns=-1,/tikz/every even column/.append style={column sep=2pt},at={(0.5,-0.78)},anchor=north},typeset ticklabels with strut}]
\input{figures/paper_palette}
\begin{axis}[name=pa,width=3.4cm,xmin=-0.5,xmax=7.5,ymin=0,ymax=0.5,ytick={0,0.25,0.5},yticklabels={0,0.25,0.5},xtick={0,1,2,3,4,5,6,7},xticklabels={\texttt{a},\texttt{two},\texttt{others},\texttt{some},\texttt{four},\texttt{another},\texttt{other},\texttt{the}},x tick label style={font=\fontsize{10}{11}\selectfont,rotate=65,anchor=north east,inner sep=1pt},xtick style={draw=none},ylabel={probability},ylabel style={yshift=-3pt},title={(a) Distributions}]
\fill[fklc,opacity=.06] (axis cs:-0.5,0) rectangle (axis cs:0.5,0.5);
\fill[fklc,opacity=.06] (axis cs:0.5,0) rectangle (axis cs:1.5,0.5);
\fill[rklc,opacity=.06] (axis cs:1.5,0) rectangle (axis cs:2.5,0.5);
\fill[rklc,opacity=.06] (axis cs:2.5,0) rectangle (axis cs:3.5,0.5);
\fill[fklc,opacity=.06] (axis cs:3.5,0) rectangle (axis cs:4.5,0.5);
\fill[rklc,opacity=.06] (axis cs:4.5,0) rectangle (axis cs:5.5,0.5);
\fill[rklc,opacity=.06] (axis cs:5.5,0) rectangle (axis cs:6.5,0.5);
\fill[rklc,opacity=.06] (axis cs:6.5,0) rectangle (axis cs:7.5,0.5);
\draw[black!22,line width=.6pt] (axis cs:0,0.4438) -- (axis cs:0,0.1385);
\draw[black!22,line width=.6pt] (axis cs:1,0.3050) -- (axis cs:1,0.1778);
\draw[black!22,line width=.6pt] (axis cs:2,0.1441) -- (axis cs:2,0.1778);
\draw[black!22,line width=.6pt] (axis cs:3,0.0874) -- (axis cs:3,0.3764);
\draw[black!22,line width=.6pt] (axis cs:4,0.0092) -- (axis cs:4,0.0000);
\draw[black!22,line width=.6pt] (axis cs:5,0.0030) -- (axis cs:5,0.1078);
\draw[black!22,line width=.6pt] (axis cs:6,0.0030) -- (axis cs:6,0.0100);
\draw[black!22,line width=.6pt] (axis cs:7,0.0018) -- (axis cs:7,0.0061);
\addplot[tch,only marks,mark=*,mark size=1.7pt,mark options={fill=tch,draw=white,line width=.3pt}] coordinates {(0,0.4438) (1,0.3050) (2,0.1441) (3,0.0874) (4,0.0092) (5,0.0030) (6,0.0030) (7,0.0018)};\addlegendentry{teacher $p$}
\addplot[stu,only marks,mark=*,mark size=1.7pt,mark options={fill=stu,draw=white,line width=.3pt}] coordinates {(0,0.1385) (1,0.1778) (2,0.1778) (3,0.3764) (4,0.0000) (5,0.1078) (6,0.0100) (7,0.0061)};\addlegendentry{student $q$}
\end{axis}
\begin{axis}[name=pb,at={(pa.east)},anchor=west,xshift=45.4pt,width=3.4cm,xmin=-0.5,xmax=8.5,ymin=60,ymax=100,ytick={60,70,80,90,100},xtick={0,1,2,3,5,6,7,8},xticklabels={2--4,4--8,8--20,$\geq$20,2--4,4--8,8--20,$\geq$20},x tick label style={font=\fontsize{10}{11}\selectfont,rotate=60,anchor=north east,inner sep=1pt},xtick style={draw=none},ylabel={Moved toward teacher (\%)},ylabel style={font=\fontsize{10}{11}\selectfont,yshift=-4pt},title={(b) Correction direction}]
\fill[fklc,opacity=.045] (axis cs:-0.5,60) rectangle (axis cs:3.5,100);
\fill[rklc,opacity=.045] (axis cs:4.5,60) rectangle (axis cs:8.5,100);
\node[font=\fontsize{10}{11}\selectfont,anchor=north] at (axis description cs:0.222,-.53) {Under, $p/q$};
\node[font=\fontsize{10}{11}\selectfont,anchor=north] at (axis description cs:0.778,-.53) {Over, $q/p$};
\addplot[fklc,mark=*,mark size=1.6pt,line width=.85pt,mark options={solid,fill=fklc,draw=white,line width=.25pt}] coordinates {(0,99.85) (1,99.71) (2,99.77) (3,99.74)};
\addlegendentry{FKL}
\addplot[fklc,mark=*,mark size=1.6pt,line width=.85pt,mark options={solid,fill=fklc,draw=white,line width=.25pt},forget plot] coordinates {(5,84.00) (6,80.18) (7,78.52) (8,75.11)};
\addplot[rklc,mark=square*,mark size=1.4pt,line width=.85pt,mark options={solid,fill=rklc,draw=white,line width=.25pt}] coordinates {(0,95.56) (1,90.67) (2,83.41) (3,70.94)};
\addlegendentry{RKL}
\addplot[rklc,mark=square*,mark size=1.4pt,line width=.85pt,mark options={solid,fill=rklc,draw=white,line width=.25pt},forget plot] coordinates {(5,79.16) (6,82.30) (7,84.50) (8,86.32)};
\addplot[dklc,mark=triangle*,mark size=1.9pt,line width=.85pt,mark options={solid,fill=dklc,draw=white,line width=.25pt}] coordinates {(0,97.41) (1,96.07) (2,94.43) (3,95.09)};
\addlegendentry{DKL}
\addplot[dklc,mark=triangle*,mark size=1.9pt,line width=.85pt,mark options={solid,fill=dklc,draw=white,line width=.25pt},forget plot] coordinates {(5,79.86) (6,82.40) (7,82.61) (8,84.74)};
\end{axis}
\pgfresetboundingbox
\path[use as bounding box] ([xshift=-1.3cm,yshift=-2.35cm]pa.south west) rectangle ([xshift=.35cm,yshift=.98cm]pb.north east);
\end{tikzpicture}%
\endgroup%
\hspace{2.2pt}\endlinechar=-1\relax\input{figures/paper_palette}\endlinechar=13\relax%
\begin{tikzpicture}[baseline=(pc.south),
  rrdnote/.style={font=\fontsize{10}{11}\selectfont, black!75, inner sep=1.5pt},
  rrdcap/.style={font=\fontsize{10}{11}\selectfont, black!80, anchor=west, inner sep=2.5pt},
]
\begin{axis}[
  name=pc, scale only axis,
  width=3.4cm, height=2.25cm,
  tick label style={font=\fontsize{10}{11}\selectfont},
  label style={font=\fontsize{10}{11}\selectfont},
  axis lines=left,
  axis line style={black!55},
  tick style={black!55},
  ymajorgrids=true,
  grid style={black!9},
  clip=false,
  xmin=0, xmax=16, xtick={0,4,8,12,16}, xticklabels={0,4K,8K,12K,16K},
  xlabel={Position in response (tokens)},
  ymin=0, ymax=25, ytick={0,5,10,15,20},
  ylabel={Repeated tokens (\%)},
  title={(c) Repetition during training},
  title style={font=\fontsize{10}{11}\selectfont\bfseries, yshift=15pt},
  every axis plot/.append style={const plot, line width=1.1pt, mark=none},
  reverse legend,
  legend style={font=\fontsize{10}{11}\selectfont, draw=none, fill=none, legend columns=2, cells={anchor=west}, /tikz/every even column/.append style={column sep=4pt}, at={(0.5,-0.78)}, anchor=north},
  legend image code/.code={\draw[#1, mark=none] (0cm,0cm) -- (0.4cm,0cm);},
]
\draw[densely dashed, black!45] (axis cs:4,0) -- ([yshift=0.10cm]axis cs:4,25);
\node[rrdnote, anchor=south east] at ([yshift=0.10cm]axis cs:4,25) {1--50};
\draw[densely dashed, black!45] (axis cs:8,0) -- ([yshift=0.10cm]axis cs:8,25);
\node[rrdnote, anchor=south] at ([yshift=0.10cm]axis cs:8,25) {51--100};
\draw[densely dashed, black!45] (axis cs:16,0) -- ([yshift=0.10cm]axis cs:16,25);
\node[rrdnote, anchor=south] at ([yshift=0.10cm]axis cs:16,25) {101+};
\addplot[rrdStepD] coordinates {(0,0.53) (4,1.31) (8,1.02) (16,1.02)};\addlegendentry{Step 200}
\addplot[rrdStepC] coordinates {(0,0.95) (4,0.70) (8,1.05) (16,1.05)};\addlegendentry{Step 100}
\addplot[rrdStepB] coordinates {(0,2.18) (4,4.25) (8,3.97) (16,3.97)};\addlegendentry{Step 50}
\addplot[rrdStepA] coordinates {(0,6.68) (4,15.65) (8,21.05) (16,21.05)};\addlegendentry{Initialization}
\node[rrdnote, anchor=south] at (axis cs:2,6.68) {6.7};
\node[rrdnote, anchor=south] at (axis cs:6,15.65) {15.6};
\node[rrdnote, anchor=south] at (axis cs:12,21.05) {21.1};
\end{axis}
\end{tikzpicture}%
}
\caption{Qwen3-1.7B W2. (a) Teacher and student probabilities at one prefix of the calibrated student. (b) Share of under- and overpredicted candidates at high-entropy prefixes that each objective moves toward the teacher. (c) Share of repeated tokens by response position during training; dashed lines mark the rollout caps, labeled with the training steps that use each cap.}
\label{fig:dynamic-kl-diagnostics}
\label{fig:rollout-length}
\end{figure}

\paragraph{Dynamic KL objective.}
At a student-generated prefix $h$, let $p_v=P(v\mid h)$ and $q_v=Q_\theta(v\mid h)$ denote the teacher and student probabilities of candidate $v$. The calibrated student exhibits two-sided probability errors: it underpredicts some candidates and overpredicts others at the same prefix (\hyperref[fig:dynamic-kl-diagnostics]{Figure~\ref*{fig:dynamic-kl-diagnostics}.a}). These errors become more frequent as student entropy rises (\hyperref[fig:opd-motivation]{Figure~\ref*{fig:opd-motivation}.c}).

Under local logit updates, FKL moves a larger fraction of severely underpredicted candidates ($p_v/q_v\geq20$) toward the teacher, whereas RKL moves a larger fraction of severely overpredicted candidates ($q_v/p_v\geq20$) (\hyperref[fig:dynamic-kl-diagnostics]{Figure~\ref*{fig:dynamic-kl-diagnostics}.b}; \hyperref[app:motivation-diagnostics]{Appendix~\ref*{app:motivation-diagnostics}}). Motivated by these complementary correction patterns, we adopt candidate-wise Dynamic KL (DKL)~\citep{DBLP:conf/emnlp/JungYKL25}, which weights the two directions according to each candidate's teacher--student probability ratio:
\begin{equation}
D_{\mathrm{dyn}}(h)=\sum_v\Big[\underbrace{\alpha_v p_v\log\frac{p_v}{q_v}}_{\text{Forward KL}}+\underbrace{(1-\alpha_v)\,q_v\log\frac{q_v}{p_v}}_{\text{Reverse KL}}\Big].
\label{eq:distillation-objective}
\end{equation}
The sum runs over vocabulary candidates $v\in\mathcal V$, and $\alpha_v=\operatorname{sg}[p_v/(p_v+q_v)]$ is the candidate-specific mixing weight computed at prefix $h$, where $\operatorname{sg}$ denotes stop-gradient. The weighting has two cases (\hyperref[fig:method-overview-placeholder]{Figure~\ref*{fig:method-overview-placeholder}}, right):
\begingroup\setlength{\topsep}{2pt}\setlength{\partopsep}{0pt}
\begin{itemize}
\setlength{\itemsep}{1pt}\setlength{\parskip}{0pt}\setlength{\parsep}{0pt}
\item \textbf{Underprediction} ($p_v>q_v$): $\alpha_v>1/2$ gives greater weight to FKL.
\item \textbf{Overprediction} ($q_v>p_v$): $\alpha_v<1/2$ gives greater weight to RKL.
\end{itemize}
\endgroup
For severely underpredicted candidates, Dynamic KL moves nearly as large a fraction toward the teacher as FKL. For severely overpredicted candidates, it approaches RKL (\hyperref[fig:dynamic-kl-diagnostics]{Figure~\ref*{fig:dynamic-kl-diagnostics}.b}). We evaluate the objective using a top-$K$ approximation $\widetilde D_{\mathrm{dyn},\theta}(h)$, with implementation details in \hyperref[app:dynamic-kl-implementation]{Appendix~\ref*{app:dynamic-kl-implementation}}.

\paragraph{On-policy optimization.}
Each step samples $B$ prompts $x_i$ from the prompt pool $\mathcal X$ and one response $y_i$ per prompt from the student. We minimize candidate-wise Dynamic KL averaged over all response tokens in the batch, at prefixes $h_{i,t}=(x_i,y_{i,<t})$:
\begin{equation}
\mathcal L_{\mathrm{OPD}}(\theta;\theta_k)=
\mathbb E_{\{x_i\}_{i=1}^{B}\sim\mathcal X,\;y_i\sim Q_{\theta_k}(\cdot\mid x_i)}\!\left[
\frac{1}{\sum_{i=1}^{B}|y_i|}
\sum_{i=1}^{B}\sum_{t=1}^{|y_i|}
\widetilde D_{\mathrm{dyn},\theta}(h_{i,t})\right].
\label{eq:opd-objective}
\end{equation}
At step $k$, we sample responses from $Q_{\theta_k}$ and hold the rollout policy, sampled responses, and teacher targets fixed while updating $\theta$~\citep{DBLP:conf/iclr/AgarwalVZSGGB24}; $\theta$ enters $\widetilde D_{\mathrm{dyn},\theta}$ only through the student probabilities. We jointly update latent weights $W$ and scales $s$~\citep{DBLP:conf/iclr/EsserMBAM20,liu2026paretoq} to minimize $\mathcal L_{\mathrm{OPD}}$. The updated student then generates fresh responses, keeping teacher supervision aligned with the prefixes visited by its evolving quantized policy.

\ifdefined\qatInlineTable\begin{table}[H]\else\ifdefined\qatTruncationOnly\begin{table}[!htbp]\else\begin{table*}[!t]\fi\fi
\centering
\ifdefined\qatTruncationOnly
\caption{Truncation rates (\%) for \hyperref[tab:main-results]{Table~\ref*{tab:main-results}}. QAT baselines and \ours{}, averaged over evaluation seeds 42, 43, and 44. Bold marks the lowest rate per model/bit group.}
\label{tab:main-results-truncation}
\else
\caption{Quantization results (\%) on Qwen3, averaged over three evaluation seeds (AutoRound: seed 42). QAT and \ours{} SDs: \hyperref[tab:qat-run-sd]{Table~\ref*{tab:qat-run-sd}}. Bold: best quantized score per model/bit group. W/A: weight/activation bits.}
\label{tab:main-results}
\label{tab:baseline-1p7b}
\label{tab:baseline-4b}
\label{tab:baseline-8b}
\fi
\begingroup
\qatTableSetup
\small
\newcommand{\secondscorecap}[2]{#1\,(#2)}
\setlength{\tabcolsep}{0pt}
\ifdefined\qatTruncationOnly
  \def\lowtr{\bfseries}
  \renewcommand{\scorecap}[2]{#2}
  \renewcommand{\bestscorecap}[2]{#2}
  \ifdefined\secondscorecap
    \renewcommand{\secondscorecap}[2]{#2}
  \fi
\else
  \def\lowtr{}
  \renewcommand{\scorecap}[2]{#1}
  \renewcommand{\bestscorecap}[2]{\textbf{#1}}
  \ifdefined\secondscorecap
    \renewcommand{\secondscorecap}[2]{#1}
  \fi
\fi

\newcommand{\pr}{\textnormal{-}}
\ifdefined\qatTruncationOnly\newcommand{\qatNonQatRows}[1]{}\else\newcommand{\qatNonQatRows}[1]{#1}\fi
\begin{tabularx}{\textwidth}{Y}
\toprule
\qatTableHead{Method} & \qatTableHead{W/A} & GSM8K & \makecell{MATH-\\500} & \makecell{AIME\\24--26} & \makecell{Human\\Eval+} & MBPP+ & \makecell{LiveCode\\Bench} & \makecell{MMLU-\\Redux} & \makecell{GPQA-\\Diamond} & IFEval & \qatTableHead{Avg.} \\
\qatTableHeaderRule
\multicolumn{12}{l}{\qatTableModel{Qwen3-1.7B}} \\
\qatNonQatRows{
Full precision & 16/16 & \scorecap{90.55}{0.08} & \scorecap{90.13}{0.40} & \scorecap{41.85}{6.30} & \scorecap{79.07}{0.00} & \scorecap{68.96}{0.88} & \scorecap{53.46}{0.92} & \scorecap{74.36}{0.04} & \scorecap{38.72}{0.17} & \scorecap{71.10}{5.11} & \scorecap{67.58}{1.54} \\
\midrule
}
\qatNonQatRows{
GPTQ & 3/16 & \scorecap{53.70}{15.92} & \scorecap{20.13}{53.27} & \scorecap{0.37}{78.89} & \scorecap{4.47}{78.25} & \scorecap{15.43}{81.04} & \scorecap{0.25}{92.51} & \scorecap{10.39}{50.46} & \scorecap{8.75}{70.71} & \scorecap{31.73}{70.73} & \scorecap{16.14}{65.75} \\
AutoRound & 3/16 & \scorecap{83.78}{0.23} & \scorecap{75.20}{5.60} & \scorecap{14.44}{20.00} & \scorecap{57.93}{3.05} & \scorecap{55.03}{1.85} & \scorecap{26.92}{18.77} & \scorecap{64.49}{0.18} & \scorecap{29.29}{1.01} & \scorecap{60.63}{7.58} & \scorecap{51.97}{6.47} \\
}
EfficientQAT & 3/16 & \scorecap{74.00}{8.82} & \scorecap{69.60}{11.80} & \scorecap{9.63}{57.78} & \scorecap{57.52}{6.10} & \scorecap{43.03}{9.26} & \scorecap{24.99}{19.12} & \scorecap{60.74}{5.02} & \scorecap{16.50}{54.21} & \scorecap{46.89}{30.44} & \scorecap{44.77}{22.50} \\
BitDistiller & 3/16 & \scorecap{80.39}{0.25} & \scorecap{66.33}{7.07} & \scorecap{8.52}{21.48} & \scorecap{56.71}{1.42} & \scorecap{49.47}{3.35} & \scorecap{20.00}{15.45} & \scorecap{61.51}{0.51} & \scorecap{24.75}{6.40} & \scorecap{58.78}{8.81} & \scorecap{47.38}{7.19} \\
ReasoningQAT & 3/16 & \scorecap{86.96}{0.03} & \scorecap{82.80}{2.00} & \scorecap{23.33}{13.33} & \scorecap{66.67}{0.61} & \scorecap{58.29}{1.59} & \scorecap{36.02}{3.38} & \scorecap{68.24}{0.12} & \scorecap{30.81}{0.67} & \lowtr\scorecap{63.34}{5.73} & \scorecap{57.38}{3.05} \\
\qatTableShade\textbf{OnQ} & 3/16 & \lowtr\bestscorecap{87.64}{0.00} & \lowtr\bestscorecap{84.73}{0.87} & \lowtr\bestscorecap{32.59}{5.56} & \lowtr\bestscorecap{71.14}{0.41} & \lowtr\bestscorecap{62.35}{1.41} & \lowtr\bestscorecap{41.77}{1.23} & \lowtr\bestscorecap{69.84}{0.05} & \lowtr\bestscorecap{30.98}{0.34} & \bestscorecap{65.87}{5.85} & \lowtr\bestscorecap{60.77}{1.75} \\
\midrule
\qatNonQatRows{
GPTQ & 2/16 & \scorecap{1.92}{99.06} & \scorecap{2.53}{99.27} & \scorecap{0.00}{99.63} & \scorecap{0.00}{99.39} & \scorecap{0.00}{99.47} & \scorecap{0.00}{98.64} & \scorecap{3.32}{98.00} & \scorecap{3.03}{97.31} & \scorecap{11.65}{98.95} & \scorecap{2.49}{98.86} \\
AutoRound & 2/16 & \scorecap{11.30}{56.94} & \scorecap{4.80}{94.00} & \scorecap{0.00}{98.89} & \scorecap{6.71}{25.61} & \scorecap{10.85}{22.75} & \scorecap{0.76}{76.30} & \scorecap{31.89}{8.00} & \scorecap{15.66}{42.42} & \scorecap{24.03}{36.97} & \scorecap{11.78}{51.32} \\
}
EfficientQAT & 2/16 & \scorecap{46.70}{16.40} & \scorecap{36.20}{32.53} & \scorecap{0.37}{81.85} & \scorecap{20.73}{25.41} & \scorecap{21.78}{23.90} & \scorecap{6.67}{60.47} & \scorecap{43.57}{8.13} & \scorecap{14.65}{54.88} & \scorecap{33.15}{48.49} & \scorecap{24.87}{39.12} \\
BitDistiller & 2/16 & \scorecap{46.63}{13.42} & \scorecap{30.33}{39.87} & \scorecap{0.00}{73.70} & \scorecap{17.48}{51.02} & \scorecap{28.75}{43.39} & \scorecap{5.43}{75.10} & \scorecap{32.39}{26.04} & \scorecap{11.62}{64.81} & \scorecap{38.20}{29.21} & \scorecap{23.42}{46.28} \\
ReasoningQAT & 2/16 & \scorecap{71.44}{0.30} & \scorecap{60.60}{8.33} & \scorecap{4.81}{28.52} & \scorecap{44.92}{4.67} & \scorecap{39.15}{4.41} & \scorecap{15.86}{24.83} & \scorecap{52.41}{0.38} & \scorecap{24.92}{4.55} & \scorecap{53.23}{9.24} & \scorecap{40.82}{9.47} \\
\qatTableShade\textbf{OnQ} & 2/16 & \lowtr\bestscorecap{79.05}{0.03} & \lowtr\bestscorecap{69.33}{2.20} & \lowtr\bestscorecap{10.37}{14.81} & \lowtr\bestscorecap{47.76}{1.42} & \lowtr\bestscorecap{42.59}{1.85} & \lowtr\bestscorecap{19.72}{4.52} & \lowtr\bestscorecap{55.49}{0.12} & \lowtr\bestscorecap{27.10}{2.02} & \lowtr\bestscorecap{55.70}{7.76} & \lowtr\bestscorecap{45.24}{3.86} \\
\qatTableModelRule
\multicolumn{12}{l}{\qatTableModel{Qwen3-4B}} \\
\qatNonQatRows{
Full precision & 16/16 & \scorecap{94.59}{0.00} & \scorecap{94.93}{0.53} & \scorecap{70.37}{8.15} & \scorecap{89.02}{0.61} & \scorecap{77.34}{0.62} & \scorecap{71.66}{0.92} & \scorecap{83.95}{0.06} & \scorecap{54.04}{0.34} & \scorecap{83.43}{1.42} & \scorecap{79.93}{1.40} \\
\midrule
}
\qatNonQatRows{
GPTQ & 3/16 & \scorecap{93.53}{0.08} & \scorecap{93.73}{0.87} & \lowtr\scorecap{51.85}{6.30} & \bestscorecap{85.98}{1.63} & \scorecap{73.02}{0.88} & \scorecap{56.97}{4.23} & \scorecap{78.65}{0.42} & \lowtr\scorecap{43.94}{0.34} & \scorecap{76.52}{3.39} & \scorecap{72.69}{2.01} \\
AutoRound & 3/16 & \scorecap{92.57}{0.00} & \scorecap{88.60}{2.80} & \scorecap{38.89}{30.00} & \scorecap{81.10}{0.61} & \scorecap{70.90}{1.85} & \scorecap{50.81}{12.51} & \scorecap{79.16}{0.07} & \scorecap{44.95}{1.01} & \scorecap{79.30}{2.59} & \scorecap{69.59}{5.72} \\
}
EfficientQAT & 3/16 & \scorecap{92.62}{0.10} & \scorecap{90.47}{1.53} & \scorecap{40.37}{13.33} & \scorecap{81.10}{2.44} & \scorecap{64.99}{3.53} & \scorecap{51.75}{9.57} & \scorecap{74.73}{4.65} & \scorecap{29.46}{50.17} & \scorecap{69.13}{14.36} & \scorecap{66.07}{11.08} \\
BitDistiller & 3/16 & \scorecap{92.67}{0.15} & \scorecap{88.93}{3.33} & \scorecap{36.67}{22.22} & \scorecap{83.74}{1.42} & \scorecap{71.34}{2.20} & \scorecap{51.22}{10.43} & \scorecap{79.42}{0.21} & \scorecap{39.90}{1.35} & \scorecap{76.65}{4.25} & \scorecap{68.95}{5.06} \\
ReasoningQAT & 3/16 & \lowtr\scorecap{93.61}{0.00} & \scorecap{92.67}{0.87} & \scorecap{52.22}{12.96} & \lowtr\scorecap{84.15}{1.02} & \scorecap{73.81}{1.59} & \scorecap{60.92}{4.52} & \scorecap{80.89}{0.12} & \lowtr\bestscorecap{50.51}{0.34} & \scorecap{79.17}{2.46} & \scorecap{74.22}{2.65} \\
\qatTableShade\textbf{OnQ} & 3/16 & \lowtr\bestscorecap{93.91}{0.00} & \lowtr\bestscorecap{93.80}{0.33} & \lowtr\bestscorecap{62.59}{8.52} & \lowtr\scorecap{84.55}{1.02} & \lowtr\bestscorecap{75.22}{0.79} & \lowtr\bestscorecap{64.55}{2.31} & \lowtr\bestscorecap{81.40}{0.06} & \lowtr\scorecap{49.16}{0.34} & \lowtr\bestscorecap{81.45}{1.91} & \lowtr\bestscorecap{76.29}{1.70} \\
\midrule
\qatNonQatRows{
GPTQ & 2/16 & \scorecap{1.74}{99.77} & \scorecap{2.13}{99.67} & \scorecap{0.37}{96.30} & \scorecap{0.00}{99.19} & \scorecap{0.00}{98.06} & \scorecap{0.00}{97.38} & \scorecap{5.17}{97.67} & \scorecap{5.89}{97.64} & \scorecap{9.37}{99.51} & \scorecap{2.74}{98.35} \\
AutoRound & 2/16 & \scorecap{32.83}{6.67} & \scorecap{15.40}{55.00} & \scorecap{0.00}{85.56} & \scorecap{14.02}{23.78} & \scorecap{20.63}{27.25} & \scorecap{2.75}{76.30} & \scorecap{45.18}{7.42} & \scorecap{18.69}{32.83} & \scorecap{27.36}{25.88} & \scorecap{19.65}{37.85} \\
}
EfficientQAT & 2/16 & \scorecap{85.90}{0.28} & \scorecap{74.93}{3.13} & \lowtr\scorecap{11.85}{17.41} & \scorecap{50.41}{15.04} & \scorecap{45.24}{12.43} & \scorecap{25.72}{35.04} & \scorecap{63.42}{1.92} & \scorecap{24.41}{27.27} & \scorecap{52.62}{25.26} & \scorecap{48.28}{15.31} \\
BitDistiller & 2/16 & \scorecap{82.76}{1.90} & \scorecap{70.60}{18.67} & \scorecap{10.37}{70.37} & \scorecap{57.32}{12.80} & \scorecap{54.59}{20.02} & \scorecap{23.29}{59.97} & \scorecap{60.86}{3.05} & \scorecap{23.23}{22.56} & \scorecap{62.42}{15.65} & \scorecap{49.49}{25.00} \\
ReasoningQAT & 2/16 & \scorecap{86.25}{0.35} & \scorecap{77.60}{8.93} & \scorecap{18.89}{43.70} & \scorecap{61.99}{8.54} & \scorecap{56.70}{13.58} & \scorecap{28.34}{39.81} & \scorecap{59.12}{0.81} & \scorecap{30.64}{7.58} & \lowtr\scorecap{60.26}{6.35} & \scorecap{53.31}{14.41} \\
\qatTableShade\textbf{OnQ} & 2/16 & \lowtr\bestscorecap{91.15}{0.00} & \lowtr\bestscorecap{87.07}{2.93} & \bestscorecap{34.81}{27.04} & \lowtr\bestscorecap{77.85}{1.22} & \lowtr\bestscorecap{65.17}{2.03} & \lowtr\bestscorecap{46.64}{8.82} & \lowtr\bestscorecap{72.23}{0.11} & \lowtr\bestscorecap{37.04}{1.01} & \bestscorecap{72.58}{6.41} & \lowtr\bestscorecap{64.95}{5.51} \\
\qatTableModelRule
\multicolumn{12}{l}{\qatTableModel{Qwen3-8B}} \\
\qatNonQatRows{
Full precision & 16/16 & \scorecap{95.73}{0.03} & \scorecap{95.53}{0.33} & \scorecap{71.85}{10.74} & \scorecap{88.21}{1.02} & \scorecap{79.28}{0.53} & \scorecap{73.59}{0.98} & \scorecap{87.42}{0.00} & \scorecap{59.93}{0.17} & \scorecap{84.84}{1.66} & \scorecap{81.82}{1.72} \\
\midrule
}
\qatNonQatRows{
GPTQ & 3/16 & \scorecap{94.92}{0.13} & \lowtr\bestscorecap{95.20}{0.27} & \lowtr\scorecap{57.41}{5.19} & \lowtr\scorecap{86.99}{0.81} & \scorecap{75.57}{0.79} & \scorecap{63.44}{3.06} & \scorecap{83.98}{0.06} & \lowtr\scorecap{50.34}{0.34} & \scorecap{80.04}{4.93} & \lowtr\scorecap{76.43}{1.73} \\
AutoRound & 3/16 & \scorecap{94.84}{0.00} & \scorecap{91.00}{3.00} & \scorecap{52.22}{24.44} & \scorecap{85.37}{1.83} & \scorecap{75.13}{1.85} & \scorecap{57.63}{9.00} & \scorecap{83.81}{0.09} & \scorecap{47.47}{1.52} & \scorecap{81.52}{2.77} & \scorecap{74.33}{4.95} \\
}
EfficientQAT & 3/16 & \scorecap{88.35}{8.47} & \scorecap{92.20}{1.47} & \lowtr\scorecap{44.81}{11.85} & \scorecap{63.62}{23.58} & \scorecap{67.81}{2.91} & \scorecap{55.92}{14.09} & \scorecap{78.61}{6.67} & \scorecap{28.96}{61.28} & \scorecap{69.38}{12.75} & \scorecap{65.52}{15.90} \\
BitDistiller & 3/16 & \lowtr\scorecap{95.00}{0.00} & \scorecap{93.67}{1.53} & \scorecap{61.85}{20.00} & \scorecap{85.57}{1.42} & \scorecap{75.75}{1.76} & \scorecap{65.31}{5.21} & \scorecap{84.60}{0.06} & \scorecap{53.03}{0.51} & \scorecap{82.07}{3.76} & \scorecap{77.43}{3.81} \\
ReasoningQAT & 3/16 & \lowtr\bestscorecap{95.27}{0.00} & \scorecap{94.47}{1.00} & \lowtr\scorecap{59.26}{11.85} & \lowtr\bestscorecap{88.01}{0.81} & \bestscorecap{76.37}{0.71} & \scorecap{65.91}{3.76} & \scorecap{85.07}{0.06} & \lowtr\scorecap{54.88}{0.34} & \scorecap{81.33}{3.76} & \scorecap{77.84}{2.48} \\
\qatTableShade\textbf{OnQ} & 3/16 & \scorecap{95.22}{0.03} & \lowtr\scorecap{94.80}{0.67} & \bestscorecap{67.04}{12.59} & \scorecap{87.20}{1.02} & \lowtr\scorecap{76.10}{0.62} & \lowtr\bestscorecap{69.23}{2.59} & \lowtr\bestscorecap{85.57}{0.03} & \lowtr\bestscorecap{55.39}{0.34} & \lowtr\bestscorecap{82.13}{3.02} & \lowtr\bestscorecap{79.19}{2.32} \\
\midrule
\qatNonQatRows{
GPTQ & 2/16 & \scorecap{1.26}{99.82} & \scorecap{1.53}{99.80} & \scorecap{0.00}{100.00} & \scorecap{0.00}{98.37} & \scorecap{0.00}{99.82} & \scorecap{0.00}{99.43} & \scorecap{6.44}{99.27} & \scorecap{6.23}{98.82} & \scorecap{10.47}{99.51} & \scorecap{2.88}{99.43} \\
AutoRound & 2/16 & \scorecap{60.88}{6.07} & \scorecap{34.00}{47.80} & \scorecap{0.00}{93.33} & \scorecap{32.32}{21.95} & \scorecap{43.92}{24.34} & \scorecap{7.39}{77.73} & \scorecap{57.40}{9.33} & \scorecap{19.19}{57.58} & \scorecap{50.09}{31.79} & \scorecap{33.91}{41.10} \\
}
EfficientQAT & 2/16 & \scorecap{81.00}{9.45} & \scorecap{75.00}{13.00} & \scorecap{11.48}{75.56} & \scorecap{67.68}{5.69} & \scorecap{54.67}{7.67} & \scorecap{33.97}{32.35} & \scorecap{66.21}{9.19} & \scorecap{18.35}{65.32} & \scorecap{59.15}{20.89} & \scorecap{51.95}{26.57} \\
BitDistiller & 2/16 & \scorecap{90.60}{0.61} & \scorecap{79.00}{13.00} & \scorecap{22.59}{63.70} & \scorecap{72.56}{4.88} & \scorecap{65.87}{6.53} & \scorecap{35.45}{37.50} & \scorecap{71.76}{1.95} & \scorecap{33.84}{10.61} & \scorecap{72.09}{9.80} & \scorecap{60.42}{16.51} \\
ReasoningQAT & 2/16 & \scorecap{90.52}{0.20} & \scorecap{85.27}{7.07} & \scorecap{28.15}{43.33} & \scorecap{72.36}{4.67} & \scorecap{63.40}{8.11} & \scorecap{40.06}{27.20} & \scorecap{69.47}{0.77} & \scorecap{30.81}{5.89} & \scorecap{62.11}{6.78} & \scorecap{60.24}{11.56} \\
\qatTableShade\textbf{OnQ} & 2/16 & \lowtr\bestscorecap{93.48}{0.00} & \lowtr\bestscorecap{90.60}{2.33} & \lowtr\bestscorecap{44.07}{23.33} & \lowtr\bestscorecap{82.93}{0.81} & \lowtr\bestscorecap{70.72}{1.23} & \lowtr\bestscorecap{55.04}{5.97} & \lowtr\bestscorecap{78.75}{0.03} & \lowtr\bestscorecap{43.43}{0.34} & \lowtr\bestscorecap{77.08}{5.42} & \lowtr\bestscorecap{70.68}{4.39} \\
\bottomrule
\end{tabularx}
\endgroup
\ifdefined\qatInlineTable\end{table}\else\ifdefined\qatTruncationOnly\end{table}\else\end{table*}\fi\fi

\paragraph{Response-length schedule.}
Right after calibration, the student repeats itself, mostly late in long responses, and the repetition fades within the first 100 training steps (\hyperref[fig:dynamic-kl-diagnostics]{Figure~\ref*{fig:dynamic-kl-diagnostics}.c}; \hyperref[app:motivation-diagnostics]{Appendix~\ref*{app:motivation-diagnostics}}). For Qwen3-1.7B at 2 bits, the share of repeated tokens at initialization rises from 6.7\% in the first 4K response tokens to 21.1\% beyond 8K; by step 100 it is about 1\% at every position. Since $\mathcal L_{\mathrm{OPD}}$ weights all response tokens equally, long early rollouts would spend supervision on this repeated text. We therefore cap responses at 4{,}096 tokens for steps 1--50, 8{,}192 for steps 51--100, and 16{,}384 afterwards for all model sizes and bit widths.

\section{Experimental Setup}
\label{sec:experiments}
\label{sec:setup}
\label{sec:results}
\paragraph{Training.}
We conduct experiments on Qwen3-1.7B, 4B, and 8B~\citep{yang2025qwen3} at 2- and 3-bit weight precision (W2 and W3). \ours{} uses uniform asymmetric weight quantization with group size 128 and BF16 activations~\citep{DBLP:conf/acl/ChenSXWGZ025}. Block-wise calibration uses 4{,}096 windows of 2{,}048 tokens, with two epochs per block. OPD starts from the calibrated checkpoint and uses a multi-domain pool of 25{,}600 prompts spanning mathematics, code, science, and general instruction following. Each student is supervised by its original BF16 counterpart. OPD uses AdamW~\citep{DBLP:conf/iclr/LoshchilovH19} at a constant learning rate of $10^{-6}$. OPD runs for 400 steps on Qwen3-1.7B and 300 steps on Qwen3-4B and 8B.

\paragraph{Baselines.}
We compare \ours{} with two groups of baselines (\hyperref[tab:main-results]{Table~\ref*{tab:main-results}}): GPTQ~\citep{frantar2022gptq} and AutoRound~\citep{DBLP:conf/emnlp/ChengZSCHLL24} for PTQ, and EfficientQAT~\citep{DBLP:conf/acl/ChenSXWGZ025}, BitDistiller~\citep{DBLP:conf/acl/DuZCGCCX24}, and ReasoningQAT~\citep{DBLP:conf/iclr/OkoshiOFM26} for QAT. Their training data and configurations are detailed in \hyperref[app:training-data]{Appendix~\ref*{app:training-data}} and \hyperref[app:baseline-configurations]{Appendix~\ref*{app:baseline-configurations}}.

\paragraph{Evaluation.}
We evaluate in thinking mode with EvalScope~\citep{evalscope_2024} on nine benchmarks: GSM8K~\citep{DBLP:journals/corr/abs-2110-14168}, MATH-500~\citep{DBLP:conf/nips/HendrycksBKABTS21,DBLP:conf/iclr/LightmanKBEBLLS24}, and AIME 24--26~\citep{aime24,aime25,aime26} for mathematics; HumanEval+, MBPP+~\citep{DBLP:journals/corr/abs-2107-03374,DBLP:journals/corr/abs-2108-07732,DBLP:conf/nips/LiuXW023}, and LiveCodeBench~\citep{DBLP:conf/iclr/JainHGLYZWSSS25} for code; MMLU-Redux~\citep{DBLP:conf/naacl/GemaLHDMSHZDMBMHKKM25} and GPQA-Diamond~\citep{rein2024gpqa} for knowledge and science; and IFEval~\citep{DBLP:journals/corr/abs-2311-07911} for instruction following. We report pass@1 for code and accuracy for the other tasks. We use a response cap of 32{,}768 tokens, with 8{,}192-token exceptions for selected PTQ settings. We count responses terminated by the output-token limit as truncated. Scores are percentages; Avg.\ is the unweighted mean across the nine benchmarks. Full-precision, GPTQ, and QAT results average evaluation seeds 42, 43, and 44; AutoRound uses seed 42. Decoding settings and benchmark versions are given in \hyperref[app:evaluation]{Appendix~\ref*{app:evaluation}}.

\section{Main Results}
\label{sec:main-results}
\begin{figure*}[!t]
\centering
\includegraphics[width=\textwidth]{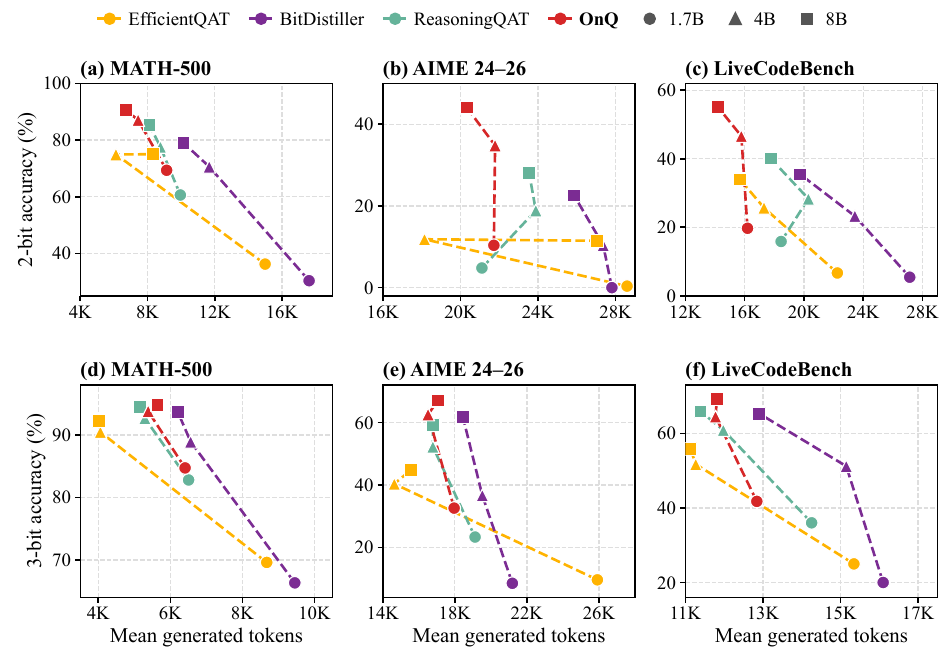}
\caption{Accuracy versus mean generated tokens for Qwen3 models at 2 bits (top, a--c) and 3 bits (bottom, d--f), averaged over three evaluation seeds. Dashed lines connect model sizes within each method.}
\label{fig:recovery-cost}
\end{figure*}

\textbf{\ours{} outperforms both PTQ and QAT baselines on every backbone and bit width.} It achieves the highest nine-benchmark average in all six model/bit settings and the highest score on 48 of the 54 benchmark--setting pairs (\hyperref[tab:main-results]{Table~\ref*{tab:main-results}}).

\textbf{The advantage of \ours{} grows as precision falls.} At every model size, the margin over the strongest baseline is larger at 2 bits than at 3 bits, and at 2 bits it exceeds 10 points on Qwen3-4B and 8B. At 2 bits, the PTQ baselines perform poorly, while \ours{}, trained on the trajectories that the 2-bit student actually generates, achieves the highest score on every benchmark at every model size. On Qwen3-4B at 2 bits, it achieves a higher nine-benchmark average than every 2-bit baseline on Qwen3-8B, a model twice its size. At 3 bits, \ours{} comes within four points of the full-precision average on Qwen3-4B and 8B.

\textbf{The largest gains occur on benchmarks that require long-horizon reasoning.} In five of the six settings, the largest margin of \ours{} over the strongest baseline falls on AIME or LiveCodeBench, and among the code benchmarks it falls on LiveCodeBench in every setting. HumanEval+ and MBPP+ ask for single functions from short specifications, whereas LiveCodeBench poses competitive-programming problems that require long-horizon reasoning before the final program; even at full precision, LiveCodeBench responses are about three times as long. Among the mathematics benchmarks, the largest margin likewise falls on AIME in every setting but one. For every QAT method and model size, AIME and LiveCodeBench elicit much longer responses than MATH-500 (\hyperref[fig:recovery-cost]{Figure~\ref*{fig:recovery-cost}}), so the student conditions on its own outputs over thousands of tokens, where small errors compound into wrong or unfinished solutions. Fixed-response QAT supervises only reference prefixes and never trains on these long student trajectories, whereas \ours{} supervises them directly.

\section{Analysis}
\label{sec:analysis}

\subsection{How Does \ours{} Change Low-Bit Generation?}

\textbf{\ours{} is more accurate with fewer generated tokens.} \hyperref[fig:recovery-cost]{Figure~\ref*{fig:recovery-cost}} plots accuracy against mean response length. At 2 bits, across MATH-500, AIME, and LiveCodeBench, \ours{} achieves the highest accuracy at all three model sizes and shorter mean responses than the most accurate baseline in eight of the nine cases. On Qwen3-4B LiveCodeBench, accuracy rises from 28.34\% for ReasoningQAT to 46.64\% while the mean response shortens by more than a fifth: \ours{} produces fewer and shorter incorrect responses. At 3 bits, \ours{} again achieves the highest accuracy among QAT methods on all three benchmarks at every model size, with shorter mean responses than the most accurate baseline in six of the nine cases (\hyperref[fig:recovery-cost]{Figure~\ref*{fig:recovery-cost}.d--f}). On Qwen3-1.7B AIME, accuracy rises from 23.33\% for ReasoningQAT to 32.59\% while the mean response shortens by 6\%. Beyond these three benchmarks, \ours{} has the lowest average truncation rate among QAT methods in all six settings (\hyperref[app:truncation-main]{Appendix~\ref*{app:truncation-main}}).

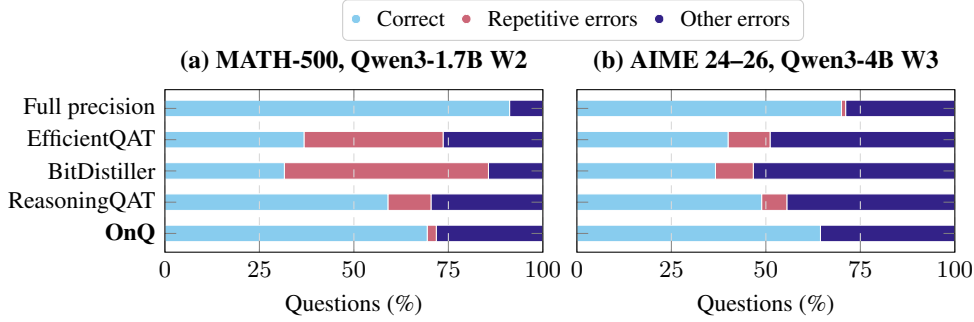
\begin{figure}[t]
\centering
\begin{tikzpicture}[/pgfplots/every axis/.append style={
 scale only axis,width=5.0cm,height=2.15cm,
 xbar stacked,bar width=6.2pt,ymin=-.6,ymax=4.6,y dir=reverse,xmin=0,xmax=100,
 ytick={0,1,2,3,4},yticklabels={Full precision,EfficientQAT,BitDistiller,ReasoningQAT,\textbf{OnQ}},
 xtick={0,25,50,75,100},tick label style={font=\fontsize{9}{10}\selectfont},
 label style={font=\fontsize{9}{10}\selectfont},title style={font=\fontsize{9.5}{11}\selectfont\bfseries,yshift=-3pt},
 xmajorgrids,grid style={black!15,dashed},axis on top,
 legend style={font=\fontsize{8.5}{10}\selectfont,draw=black!20,rounded corners=2pt,legend columns=3,column sep=4pt},
 legend image code/.code={\draw[#1] (0cm,-0.05cm) rectangle (0.13cm,0.08cm);},
 every axis plot/.append style={draw=white,line width=.35pt,mark=none}}]
\definecolor{outCorrect}{HTML}{88CCEE}
\definecolor{outRepetitive}{HTML}{CC6677}
\definecolor{outOther}{HTML}{332288}
\begin{axis}[name=oa,title={(a) MATH-500, Qwen3-1.7B W2},xlabel={Questions (\%)},legend style={at={(1.08,1.30)},anchor=south}]
\addplot[fill=outCorrect] coordinates {(91.2,0) (36.8,1) (31.6,2) (59.0,3) (69.4,4)};
\addlegendentry{Correct}
\addplot[fill=outRepetitive] coordinates {(0.0,0) (36.8,1) (54.0,2) (11.4,3) (2.4,4)};
\addlegendentry{Repetitive errors}
\addplot[fill=outOther] coordinates {(8.8,0) (26.4,1) (14.4,2) (29.6,3) (28.2,4)};
\addlegendentry{Other errors}
\end{axis}
\begin{axis}[name=ob,at={(oa.east)},anchor=west,xshift=.45cm,yticklabels=\empty,title={(b) AIME 24--26, Qwen3-4B W3},xlabel={Questions (\%)}]
\addplot[fill=outCorrect] coordinates {(70.0,0) (40.0,1) (36.6667,2) (48.8889,3) (64.4444,4)};
\addplot[fill=outRepetitive] coordinates {(1.1111,0) (11.1111,1) (10.0,2) (6.6667,3) (0.0,4)};
\addplot[fill=outOther] coordinates {(28.8889,0) (48.8889,1) (53.3333,2) (44.4444,3) (35.5556,4)};
\end{axis}
\end{tikzpicture}
\caption{Generation outcomes on MATH-500 (Qwen3-1.7B W2, a) and AIME (Qwen3-4B W3, b). \ours{} reduces repetitive errors on both benchmarks.}
\label{fig:generation-outcomes}
\end{figure}
\textbf{\ours{} substantially reduces repetitive generation in low-bit models.}
\hyperref[fig:generation-outcomes]{Figure~\ref*{fig:generation-outcomes}} separates incorrect responses into repetitive errors, in which responses emit the same sentence, block, or computation again and again, and all other errors (\hyperref[app:figure1-details]{Appendix~\ref*{app:figure1-details}}), for Qwen3-1.7B at 2 bits on MATH-500 and Qwen3-4B at 3 bits on AIME. The full-precision models produce no repetitive errors on MATH-500 and one on AIME, whereas these errors become much more frequent after quantization. The fixed-response QAT baselines produce repetitive errors on 11.4--54.0\% of the MATH-500 questions and on 6.7--11.1\% of the AIME questions. Trained on its own generations, \ours{} produces repetitive errors on only 2.4\% of the MATH-500 questions, and this reduction accounts for most of its accuracy gain over ReasoningQAT. On AIME, \ours{} produces no repetitive errors. Repetition likewise appears in the calibrated student's early rollouts and fades within the first 100 steps of on-policy training (\hyperref[sec:distillation-objective]{Section~\ref*{sec:distillation-objective}}).

\ifdefined\qatInlineTable\begin{table}[H]\else\ifdefined\qatTruncationOnly\begin{table}[!htbp]\else\begin{table*}[!t]\fi\fi
\centering
\ifdefined\qatTruncationOnly
\caption{Truncation rates (\%) for \hyperref[tab:opd-gain]{Table~\ref*{tab:opd-gain}}; bold marks the lowest rate per model/bit group.}
\label{tab:opd-gain-truncation}
\else
\caption{Scores (\%) across recovery stages. OnQ denotes calibration followed by on-policy distillation. OnQ results average evaluation seeds 42, 43, and 44; RTN and calibration results use seed 42. Bold marks the best score within each model/bit group.}
\label{tab:opd-gain}
\label{tab:rtn-start-selected}
\label{tab:rtn-start8k}
\label{tab:rtn-start32k}
\fi
\begingroup
\qatTableSetup
\small
\setlength{\tabcolsep}{0pt}
\ifdefined\qatTruncationOnly
  \def\lowtr{\bfseries}
  \renewcommand{\scorecap}[2]{#2}
  \renewcommand{\bestscorecap}[2]{#2}
  \ifdefined\secondscorecap
    \renewcommand{\secondscorecap}[2]{#2}
  \fi
\else
  \def\lowtr{}
  \renewcommand{\scorecap}[2]{#1}
  \renewcommand{\bestscorecap}[2]{\textbf{#1}}
  \ifdefined\secondscorecap
    \renewcommand{\secondscorecap}[2]{#1}
  \fi
\fi

\newcommand{\pr}{\textcolor{red}{--}}
\begin{tabularx}{\textwidth}{Y}
\toprule
\qatTableHead{Method} & \qatTableHead{W/A} & GSM8K & \makecell{MATH-\\500} & \makecell{AIME\\24--26} & \makecell{Human\\Eval+} & MBPP+ & \makecell{LiveCode\\Bench} & \makecell{MMLU-\\Redux} & \makecell{GPQA-\\Diamond} & IFEval & \qatTableHead{Avg.} \\
\qatTableHeaderRule
\multicolumn{12}{l}{\qatTableModel{Qwen3-1.7B}} \\
RTN & 3/16 & \scorecap{0.68}{94.16} & \scorecap{1.20}{94.00} & \scorecap{1.11}{97.78} & \scorecap{0.00}{98.78} & \scorecap{0.00}{97.09} & \scorecap{0.00}{93.55} & \scorecap{8.72}{99.46} & \scorecap{9.60}{100.00} & \scorecap{12.01}{86.88} & \scorecap{3.70}{95.74} \\
$+$ Calibration & 3/16 & \scorecap{87.11}{0.08} & \scorecap{80.00}{2.60} & \scorecap{17.78}{16.67} & \scorecap{58.54}{0.61} & \lowtr\scorecap{58.20}{1.06} & \scorecap{30.71}{8.34} & \scorecap{67.16}{0.12} & \scorecap{29.80}{0.51} & \scorecap{61.92}{7.21} & \scorecap{54.58}{4.13} \\
\qatTableShade\textbf{OnQ} & 3/16 & \lowtr\bestscorecap{87.64}{0.00} & \lowtr\bestscorecap{84.73}{0.87} & \lowtr\bestscorecap{32.59}{5.56} & \lowtr\bestscorecap{71.14}{0.41} & \bestscorecap{62.35}{1.41} & \lowtr\bestscorecap{41.77}{1.23} & \lowtr\bestscorecap{69.84}{0.05} & \lowtr\bestscorecap{30.98}{0.34} & \lowtr\bestscorecap{65.87}{5.85} & \lowtr\bestscorecap{60.77}{1.75} \\
\midrule
RTN & 2/16 & \scorecap{1.67}{99.85} & \scorecap{2.40}{100.00} & \scorecap{0.00}{100.00} & \scorecap{0.00}{100.00} & \scorecap{0.00}{100.00} & \scorecap{0.00}{99.91} & \scorecap{8.60}{100.00} & \scorecap{11.62}{100.00} & \scorecap{10.91}{100.00} & \scorecap{3.91}{99.97} \\
$+$ Calibration & 2/16 & \scorecap{42.84}{0.68} & \scorecap{25.60}{20.80} & \scorecap{0.00}{52.22} & \scorecap{15.85}{17.68} & \scorecap{19.84}{25.66} & \scorecap{2.84}{58.96} & \scorecap{45.07}{3.67} & \scorecap{21.72}{22.73} & \scorecap{36.41}{36.04} & \scorecap{23.35}{26.49} \\
\qatTableShade\textbf{OnQ} & 2/16 & \lowtr\bestscorecap{79.05}{0.03} & \lowtr\bestscorecap{69.33}{2.20} & \lowtr\bestscorecap{10.37}{14.81} & \lowtr\bestscorecap{47.76}{1.42} & \lowtr\bestscorecap{42.59}{1.85} & \lowtr\bestscorecap{19.72}{4.52} & \lowtr\bestscorecap{55.49}{0.12} & \lowtr\bestscorecap{27.10}{2.02} & \lowtr\bestscorecap{55.70}{7.76} & \lowtr\bestscorecap{45.24}{3.86} \\
\qatTableModelRule
\multicolumn{12}{l}{\qatTableModel{Qwen3-4B}} \\
RTN & 3/16 & \scorecap{81.58}{1.21} & \scorecap{65.00}{21.60} & \scorecap{5.56}{61.11} & \scorecap{46.95}{39.63} & \scorecap{45.77}{32.80} & \scorecap{14.22}{69.00} & \scorecap{65.26}{6.54} & \scorecap{25.25}{21.21} & \scorecap{56.56}{11.09} & \scorecap{45.13}{29.36} \\
$+$ Calibration & 3/16 & \scorecap{92.72}{0.08} & \scorecap{92.00}{1.20} & \scorecap{50.00}{21.11} & \scorecap{82.93}{1.22} & \scorecap{72.22}{1.59} & \scorecap{56.87}{6.92} & \scorecap{80.49}{0.19} & \scorecap{48.99}{0.51} & \scorecap{78.19}{3.14} & \scorecap{72.71}{3.99} \\
\qatTableShade\textbf{OnQ} & 3/16 & \lowtr\bestscorecap{93.91}{0.00} & \lowtr\bestscorecap{93.80}{0.33} & \lowtr\bestscorecap{62.59}{8.52} & \lowtr\bestscorecap{84.55}{1.02} & \lowtr\bestscorecap{75.22}{0.79} & \lowtr\bestscorecap{64.55}{2.31} & \lowtr\bestscorecap{81.40}{0.06} & \lowtr\bestscorecap{49.16}{0.34} & \lowtr\bestscorecap{81.45}{1.91} & \lowtr\bestscorecap{76.29}{1.70} \\
\midrule
RTN & 2/16 & \scorecap{1.52}{99.77} & \scorecap{2.00}{99.80} & \scorecap{0.00}{100.00} & \scorecap{0.00}{100.00} & \scorecap{0.00}{100.00} & \scorecap{0.00}{100.00} & \scorecap{1.74}{99.96} & \scorecap{2.02}{100.00} & \scorecap{12.75}{100.00} & \scorecap{2.23}{99.95} \\
$+$ Calibration & 2/16 & \scorecap{78.54}{0.61} & \scorecap{59.40}{22.20} & \scorecap{1.11}{77.78} & \scorecap{42.68}{12.80} & \scorecap{48.68}{12.43} & \scorecap{16.30}{53.84} & \scorecap{63.09}{1.72} & \scorecap{26.26}{16.16} & \scorecap{50.83}{11.09} & \scorecap{42.99}{23.18} \\
\qatTableShade\textbf{OnQ} & 2/16 & \lowtr\bestscorecap{91.15}{0.00} & \lowtr\bestscorecap{87.07}{2.93} & \lowtr\bestscorecap{34.81}{27.04} & \lowtr\bestscorecap{77.85}{1.22} & \lowtr\bestscorecap{65.17}{2.03} & \lowtr\bestscorecap{46.64}{8.82} & \lowtr\bestscorecap{72.23}{0.11} & \lowtr\bestscorecap{37.04}{1.01} & \lowtr\bestscorecap{72.58}{6.41} & \lowtr\bestscorecap{64.95}{5.51} \\
\qatTableModelRule
\multicolumn{12}{l}{\qatTableModel{Qwen3-8B}} \\
RTN & 3/16 & \scorecap{33.59}{55.12} & \scorecap{17.00}{80.80} & \scorecap{0.00}{94.44} & \scorecap{11.59}{64.02} & \scorecap{26.46}{41.27} & \scorecap{0.28}{78.29} & \scorecap{25.84}{50.56} & \scorecap{13.13}{82.83} & \scorecap{37.71}{65.43} & \scorecap{18.40}{68.09} \\
$+$ Calibration & 3/16 & \scorecap{94.54}{0.08} & \scorecap{93.00}{1.80} & \scorecap{53.33}{18.89} & \scorecap{85.37}{1.22} & \scorecap{74.60}{1.85} & \scorecap{61.33}{8.34} & \scorecap{84.77}{0.05} & \lowtr\scorecap{53.03}{0.00} & \scorecap{80.04}{4.07} & \scorecap{75.56}{4.03} \\
\qatTableShade\textbf{OnQ} & 3/16 & \lowtr\bestscorecap{95.22}{0.03} & \lowtr\bestscorecap{94.80}{0.67} & \lowtr\bestscorecap{67.04}{12.59} & \lowtr\bestscorecap{87.20}{1.02} & \lowtr\bestscorecap{76.10}{0.62} & \lowtr\bestscorecap{69.23}{2.59} & \lowtr\bestscorecap{85.57}{0.03} & \bestscorecap{55.39}{0.34} & \lowtr\bestscorecap{82.13}{3.02} & \lowtr\bestscorecap{79.19}{2.32} \\
\midrule
RTN & 2/16 & \scorecap{1.82}{98.18} & \scorecap{1.40}{98.80} & \scorecap{0.00}{98.89} & \scorecap{0.00}{98.17} & \scorecap{0.00}{99.21} & \scorecap{0.00}{99.72} & \scorecap{4.18}{97.93} & \scorecap{3.03}{98.48} & \scorecap{8.32}{99.45} & \scorecap{2.08}{98.76} \\
$+$ Calibration & 2/16 & \scorecap{84.91}{1.06} & \scorecap{65.00}{24.40} & \scorecap{15.56}{70.00} & \scorecap{56.71}{9.15} & \scorecap{58.47}{11.64} & \scorecap{18.77}{54.79} & \scorecap{69.70}{4.67} & \scorecap{27.78}{18.69} & \scorecap{53.97}{20.33} & \scorecap{50.10}{23.86} \\
\qatTableShade\textbf{OnQ} & 2/16 & \lowtr\bestscorecap{93.48}{0.00} & \lowtr\bestscorecap{90.60}{2.33} & \lowtr\bestscorecap{44.07}{23.33} & \lowtr\bestscorecap{82.93}{0.81} & \lowtr\bestscorecap{70.72}{1.23} & \lowtr\bestscorecap{55.04}{5.97} & \lowtr\bestscorecap{78.75}{0.03} & \lowtr\bestscorecap{43.43}{0.34} & \lowtr\bestscorecap{77.08}{5.42} & \lowtr\bestscorecap{70.68}{4.39} \\
\bottomrule
\end{tabularx}
\endgroup
\ifdefined\qatInlineTable\end{table}\else\ifdefined\qatTruncationOnly\end{table}\else\end{table*}\fi\fi

\subsection{Which Design Choices Drive the Gains of \ours{}?}
\label{sec:ablations}

\begin{wraptable}{r}{0.49\textwidth}
\centering
\caption{Fixed-response versus on-policy distillation on Qwen3-1.7B W2 from shared calibration. Scores in \%; best scores in bold.}
\label{tab:offline-online-token-matched}
\begingroup
\qatTableSetup
\small
\ifdefined\qatTruncationOnly
  \def\lowtr{\bfseries}
  \renewcommand{\scorecap}[2]{#2}
  \renewcommand{\bestscorecap}[2]{#2}
  \ifdefined\secondscorecap
    \renewcommand{\secondscorecap}[2]{#2}
  \fi
\else
  \def\lowtr{}
  \renewcommand{\scorecap}[2]{#1}
  \renewcommand{\bestscorecap}[2]{\textbf{#1}}
  \ifdefined\secondscorecap
    \renewcommand{\secondscorecap}[2]{#1}
  \fi
\fi

\setlength{\tabcolsep}{1pt}
\begin{tabularx}{\linewidth}{@{}l*{3}{>{\centering\arraybackslash}X}@{}}
\toprule
\qatTableHead{Method} & \qatTableHead{\makecell{MATH-\\500}} & \qatTableHead{\makecell{Human\\Eval+}} & \qatTableHead{\makecell{MMLU-\\Redux}} \\
\qatTableHeaderRule
Fixed-response FKL & \scorecap{59.00}{7.20} & \scorecap{39.63}{2.44} & \scorecap{52.28}{0.25} \\
Fixed-response DKL & \scorecap{61.00}{12.00} & \scorecap{42.07}{7.32} & \scorecap{52.68}{0.75} \\
\qatTableShade\textbf{On-policy DKL (OnQ)} & \bestscorecap{69.40}{2.00} & \bestscorecap{48.78}{2.44} & \bestscorecap{55.75}{0.18} \\
\bottomrule
\end{tabularx}
\endgroup
\end{wraptable}

\paragraph{Contribution of each stage.}
\label{sec:ablation-recovery-stages}

To separate the two stages of \ours{}, \hyperref[tab:opd-gain]{Table~\ref*{tab:opd-gain}} reports each model after RTN, after calibration, and after on-policy distillation. Direct RTN severely degrades 2-bit accuracy, motivating restoration of a working policy before training on the student's own rollouts. Calibration provides one but still struggles on long-horizon reasoning at 2 bits, as its LiveCodeBench scores show. On-policy distillation then closes about half or more of the remaining gap to full precision in every setting, with substantial gains on MATH-500 and LiveCodeBench, and lowers truncation in all six settings (\hyperref[app:truncation-stages]{Appendix~\ref*{app:truncation-stages}}).

\paragraph{On-policy versus fixed-response distillation.}
\label{sec:ablation-trajectory-source}
We compare \ours{} with fixed-response distillation using either forward KL (FKL) or Dynamic KL (DKL) on Qwen3-1.7B W2 (\hyperref[tab:offline-online-token-matched]{Table~\ref*{tab:offline-online-token-matched}}). All three runs start from the same calibrated checkpoint and use the same number of response tokens; fixed-response training uses the QAT baselines' question--response pairs and on-policy training uses our prompt pool (\hyperref[app:training-data]{Appendices~\ref*{app:training-data}} and \hyperref[app:ablation-trajectories]{\ref*{app:ablation-trajectories}}). On-policy Dynamic KL exceeds fixed-response Dynamic KL on all three benchmarks, by up to 8.40 points, and replacing FKL with Dynamic KL in fixed-response distillation adds up to 2.44.

\begin{wraptable}{r}{0.55\textwidth}
\centering
\caption{On-policy objectives on Qwen3-1.7B W2 (400 steps from a shared cold start). Scores in \%; best scores in bold.}
\label{tab:on-policy-objectives}
\begingroup
\qatTableSetup
\small
\ifdefined\qatTruncationOnly
  \def\lowtr{\bfseries}
  \renewcommand{\scorecap}[2]{#2}
  \renewcommand{\bestscorecap}[2]{#2}
  \ifdefined\secondscorecap
    \renewcommand{\secondscorecap}[2]{#2}
  \fi
\else
  \def\lowtr{}
  \renewcommand{\scorecap}[2]{#1}
  \renewcommand{\bestscorecap}[2]{\textbf{#1}}
  \ifdefined\secondscorecap
    \renewcommand{\secondscorecap}[2]{#1}
  \fi
\fi

\setlength{\tabcolsep}{1pt}
\begin{tabularx}{\linewidth}{@{}l<{\hspace{5pt}}l*{3}{>{\centering\arraybackslash}X}@{}}
\toprule
\qatTableHead{Method} & \qatTableHead{Objective} & \qatTableHead{\makecell{MATH-\\500}} & \qatTableHead{\makecell{Human\\Eval+}} & \qatTableHead{\makecell{MMLU-\\Redux}} \\
\qatTableHeaderRule
Cold start & -- & \scorecap{41.40}{76.80} & \scorecap{3.66}{17.68} & \scorecap{45.12}{4.47} \\
\midrule
Vanilla OPD & RKL & \scorecap{63.00}{8.40} & \scorecap{10.98}{5.49} & \scorecap{48.18}{0.70} \\
OPD + GRPO & RKL + RLVR & \scorecap{66.00}{7.00} & \scorecap{4.27}{10.98} & \scorecap{48.84}{0.67} \\
GKD & FKL & \scorecap{66.20}{2.00} & \bestscorecap{11.59}{4.27} & \scorecap{48.82}{0.26} \\
GRPO & RLVR & \scorecap{31.40}{100.00} & \scorecap{0.61}{92.68} & \scorecap{34.05}{97.88} \\
\qatTableShade\textbf{OnQ} & Dynamic KL & \bestscorecap{67.20}{5.60} & \scorecap{10.98}{2.44} & \bestscorecap{49.58}{0.30} \\
\bottomrule
\end{tabularx}
\endgroup
\end{wraptable}

\paragraph{Dynamic KL versus other on-policy objectives.}
\label{sec:ablation-objective}
We compare on-policy objectives using a shared initialization, training set, and step budget. \hyperref[tab:on-policy-objectives]{Table~\ref*{tab:on-policy-objectives}} reports results for \ours{}, Vanilla OPD~\citep{lu2025opd}, GKD~\citep{DBLP:conf/iclr/AgarwalVZSGGB24}, GRPO~\citep{shao2024deepseekmath}, and OPD~+~GRPO on Qwen3-1.7B W2. Vanilla OPD uses RKL, and we use the FKL variant of GKD as the other single-direction on-policy baseline. GRPO uses reinforcement learning with verifiable rewards (RLVR) alone, and OPD~+~GRPO adds this RLVR objective to sampled-RKL OPD. All runs start from a shared cold-start checkpoint and train on DeepScaleR mathematics problems~\citep{deepscaler2025} (\hyperref[app:ablation-objectives]{Appendix~\ref*{app:ablation-objectives}}).

Dynamic KL achieves the highest average over the three benchmarks and the highest MATH-500 and MMLU-Redux scores. Relative to FKL (GKD) and RKL (Vanilla OPD), it improves MATH-500 by 1.0 and 4.2 points, respectively. Relying solely on teacher supervision and trained only on mathematics prompts, \ours{} outperforms OPD~+~GRPO and GRPO on all three benchmarks, showing stronger transfer to code generation and knowledge.

\section{Conclusion}
We present \ours{}, a low-bit recovery pipeline that addresses trajectory divergence with on-policy distillation and distributional misalignment with candidate-wise Dynamic KL. On Qwen3-1.7B, 4B, and 8B at 2 and 3 bits, \ours{} achieves the highest nine-benchmark average in all six settings, with the largest gains at 2 bits and on long-horizon reasoning benchmarks. The recovered models also produce shorter responses with fewer repetitive errors. On-policy training outperforms fixed-response distillation, and Dynamic KL achieves the best average among the on-policy objectives we compare. These results establish training on self-generated trajectories as an effective route to recovering reasoning under aggressive quantization.

\bibliography{references}
\bibliographystyle{iclr2027_conference}

\appendix
\section{Experimental Details}
\label{app:experimental-details}
\subsection{Data}
\label{app:training-data}
\hyperref[tab:data]{Table~\ref*{tab:data}} lists the calibration, OPD, and QAT baseline data, drawn from OpenMathReasoning~\citep{DBLP:journals/corr/abs-2504-16891}, OpenCodeReasoning-2~\citep{DBLP:journals/corr/abs-2507-09075}, OpenScienceReasoning-2~\citep{nvidia2025osr2}, Dolci-Think~\citep{olmo2025olmo}, T\"ulu~3~\citep{lambert2025tulu}, IF-multi-constraints~\citep{DBLP:conf/nips/PyatkinMGIHDLH25}, and Nemotron-Instruction-Following-Chat~\citep{nvidia2025nemotronifchat}.

\begin{table}[!htbp]
\centering
\caption{Training data: calibration windows (2{,}048 tokens), OPD prompts, and question--response pairs shared by the three QAT baselines.}
\label{tab:data}
\begingroup
\qatTableSetup
\setlength{\tabcolsep}{0pt}
\begin{tabularx}{\textwidth}{@{}l*{3}{>{\raggedleft\arraybackslash}X}@{}}
\toprule
\qatTableHead{Source} & \qatTableHead{Calib.} & \qatTableHead{OPD} & \qatTableHead{Baselines} \\
\qatTableHeaderRule
OpenMathReasoning & 1{,}229 & 7{,}680 & 9{,}831 \\
OpenCodeReasoning-2 & 1{,}229 & 7{,}680 & 9{,}831 \\
OpenScienceReasoning-2 & 614 & 3{,}840 & 4{,}915 \\
Dolci-Think & 1{,}024 & -- & 6{,}553 \\
T\"ulu 3 & -- & 5{,}120 & -- \\
IF-multi-constraints & -- & 1{,}280 & -- \\
Nemotron-IF-Chat & -- & -- & 1{,}638 \\
\midrule
\qatTableShade\textbf{Total} & 4{,}096 & 25{,}600 & 32{,}768 \\
\bottomrule
\end{tabularx}
\endgroup
\end{table}

\subsection{Training Details}
\label{app:quantization-training}
We quantize all linear layers in the transformer blocks, with a straight-through estimator for rounding~\citep{DBLP:journals/corr/BengioLC13}. Calibration uses batch size 2 and initializes quantization ranges at 0.7 times each group's minimum and maximum weights. Learning rates start at $2\times10^{-5}$ (2-bit) or $10^{-5}$ (3-bit) for latent weights and $10^{-4}$ for scales and zero points, with cosine decay to 5\% of their initial values.

OPD uses verl~\citep{DBLP:conf/eurosys/ShengZYWZZPL025} on NVIDIA H100 and H200 GPUs, with the full-precision teacher served by vLLM~\citep{DBLP:conf/sosp/KwonLZ0ZY0ZS23}. Each step samples one response for each of 128 prompts at temperature 1.0, without top-$p$ or top-$k$ truncation, and applies a single update. AdamW uses no warmup, weight decay 0.01, and gradient-norm clipping at 1.0.

\subsection{Dynamic KL Implementation}
\label{app:dynamic-kl-implementation}
At a prefix $h$, let $\mathcal U(h)$ be the union of the teacher's and the student's top-20 tokens, and $\mathcal S(h)=\mathcal U(h)\cup\{\bot\}$, where the residual category $\bot$ carries the remaining mass $p_\bot=1-\sum_{v\in\mathcal U(h)}p_v$ and $q_\bot=1-\sum_{v\in\mathcal U(h)}q_v$. The teacher returns its probability for every token in $\mathcal U(h)$. We smooth the teacher categories as $\bar p_c=(p_c+\epsilon)/(1+|\mathcal S(h)|\epsilon)$ with $\epsilon=10^{-8}$, leave the student categories unsmoothed, and evaluate \hyperref[eq:distillation-objective]{Equation~\ref*{eq:distillation-objective}} and its mixing weights over $\mathcal S(h)$, which gives $\widetilde D_{\mathrm{dyn},\theta}(h)$. All student categories, including $q_\bot$, keep their gradients.

\subsection{Motivation Diagnostics}
\label{app:motivation-diagnostics}
\begingroup\setlength{\topsep}{2pt}\setlength{\partopsep}{0pt}\setlength{\leftmargini}{1.1em}
\begin{itemize}
\setlength{\itemsep}{1pt}\setlength{\parskip}{0pt}\setlength{\parsep}{0pt}
\item \textbf{\hyperref[fig:opd-motivation]{Figure~\ref*{fig:opd-motivation}.a--b}.} Seed-42 MATH-500 responses of Qwen3-1.7B W2 (32{,}768-token cap), grouped by the length of the full-precision response.
\item \textbf{\hyperref[fig:opd-motivation]{Figure~\ref*{fig:opd-motivation}.c} and \hyperref[fig:dynamic-kl-diagnostics]{Figure~\ref*{fig:dynamic-kl-diagnostics}.a--b}.} We compare teacher and student probabilities at 16{,}384 evenly spaced prefixes of 128 calibrated-student responses (temperature 1, 2{,}048-token cap).
\item \textbf{Under- and overpredicted.} A candidate is underpredicted when $q_v<p_v$ and overpredicted when $q_v>p_v$. We count it only if the two probabilities differ by at least 0.05 and by a factor of at least 2; the residual category is excluded.
\item \textbf{\hyperref[fig:opd-motivation]{Figure~\ref*{fig:opd-motivation}.c}.} For each student-entropy bin, the share of prefixes with an underpredicted candidate, an overpredicted candidate, or both. Bands are 95\% bootstrap intervals over prompts.
\item \textbf{\hyperref[fig:dynamic-kl-diagnostics]{Figure~\ref*{fig:dynamic-kl-diagnostics}.a}.} The eight candidates most probable under the teacher or the student at one prefix.
\item \textbf{\hyperref[fig:dynamic-kl-diagnostics]{Figure~\ref*{fig:dynamic-kl-diagnostics}.b}.} For each objective, we check whether one small gradient step on the student logits raises the probability of an underpredicted candidate or lowers that of an overpredicted one. We use prefixes with entropy of at least 0.5 nats. Among the 1{,}913 underpredicted candidates off by a factor of at least 20, FKL does so for 99.7\%, RKL for 70.9\%, and Dynamic KL for 95.1\%. Among the 1{,}828 overpredicted ones, FKL does so for 75.1\%, RKL for 86.3\%, and Dynamic KL for 84.7\%.
\item \textbf{\hyperref[fig:dynamic-kl-diagnostics]{Figure~\ref*{fig:dynamic-kl-diagnostics}.c}.} Responses to 128 training prompts (16{,}384-token cap) from the checkpoints at initialization and steps 50, 100, and 200. A token is repeated if it lies in the third or later non-overlapping occurrence of an identical 32-token span.
\end{itemize}
\endgroup

\subsection{Baseline Configurations}
\label{app:baseline-configurations}
\hyperref[tab:baseline-configurations]{Table~\ref*{tab:baseline-configurations}} summarizes the optimization settings.
\begingroup\setlength{\topsep}{2pt}\setlength{\partopsep}{0pt}\setlength{\leftmargini}{1.1em}
\begin{itemize}
\setlength{\itemsep}{1pt}\setlength{\parskip}{0pt}\setlength{\parsep}{0pt}
\item \textbf{Training.} The three QAT baselines use the 32{,}768 single-turn question--response pairs in \hyperref[tab:data]{Table~\ref*{tab:data}}, with sequences cut at 8{,}192 tokens.
\item \textbf{EfficientQAT.} Starts from the same calibrated checkpoint as \ours{} and follows its end-to-end stage settings.
\item \textbf{ReasoningQAT.} Starts from the same calibrated checkpoint and follows its released implementation.
\item \textbf{BitDistiller.} Follows its original settings, with its own clipping initialization and CAKLD objective.
\end{itemize}
\endgroup

\begin{table}[!htbp]
\centering
\caption{Baseline configurations for \hyperref[tab:main-results]{Table~\ref*{tab:main-results}}. Batch size counts sequences per optimizer step. All methods use group size 128 and 2- or 3-bit weights.}
\label{tab:baseline-configurations}
\begingroup
\qatTableSetup
\setlength{\tabcolsep}{5pt}
\begin{tabularx}{\textwidth}{@{}llrr>{\raggedright\arraybackslash}X@{}}
\toprule
\textbf{Method} & \textbf{Setting} & \textbf{Steps} & \textbf{Batch} & \textbf{Learning rate} \\
\midrule
EfficientQAT & W2 / W3 & 1{,}024 & 32 & $2\!\times\!10^{-5}$ / $10^{-5}$ \\
BitDistiller & W2 / W3 & 768 / 256 & 128 & $8\!\times\!10^{-6}$ \\
ReasoningQAT & 1.7B W2 & 1{,}536 & 64 & $5\!\times\!10^{-6}$ \\
 & 4B/8B W2 & 512 & 64 & $10^{-4}$ \\
 & All W3 & 512 & 64 & $10^{-6}$ \\
\midrule
\multicolumn{5}{@{}l}{\textbf{PTQ calibration:} 4{,}096 sequences of 2{,}048 tokens, seed 42.} \\
\midrule
GPTQ & \multicolumn{4}{p{.79\textwidth}@{}}{Asymmetric; block size 128; damping 0.01; activation ordering disabled.} \\
AutoRound & \multicolumn{4}{p{.79\textwidth}@{}}{Symmetric; 200 iterations per block; batch size 8; rounding and min/max learning rates 0.005.} \\
\bottomrule
\end{tabularx}
\endgroup
\end{table}

\subsection{Evaluation}
\label{app:evaluation}
Decoding uses temperature 0.6, top-$p$ 1.0, and top-$k$ 20, with one response per question. We use EvalScope 1.5.1, with EvalPlus 0.3.1 for the full HumanEval+ and MBPP+ tests, and report prompt-level strict accuracy for IFEval. AIME 24--26 comprises the 90 AIME I and II problems from 2024 to 2026, and LiveCodeBench uses the release-v6 set of 1{,}055 problems~\citep{DBLP:conf/iclr/JainHGLYZWSSS25}.

\subsection{Ablation Settings}
\label{app:ablation-settings}
\label{app:ablation-trajectories}
\label{app:ablation-objectives}
\begingroup\setlength{\topsep}{2pt}\setlength{\partopsep}{0pt}\setlength{\leftmargini}{1.1em}
\begin{itemize}
\setlength{\itemsep}{1pt}\setlength{\parskip}{0pt}\setlength{\parsep}{0pt}
\item \textbf{On-policy versus fixed-response (\hyperref[tab:offline-online-token-matched]{Table~\ref*{tab:offline-online-token-matched}}).} All runs use the same number of response tokens in the loss; fixed-response sequences are cut at 8{,}192 tokens as for the QAT baselines.
\item \textbf{On-policy objectives (\hyperref[tab:on-policy-objectives]{Table~\ref*{tab:on-policy-objectives}}).} All runs train on DeepScaleR mathematics problems because RLVR requires verifiable answers. They start from a shared cold-start checkpoint and train for 400 steps.
\end{itemize}
\endgroup

\section{Response Classification}
\label{app:figure1-details}
\begin{figure}[!tp]
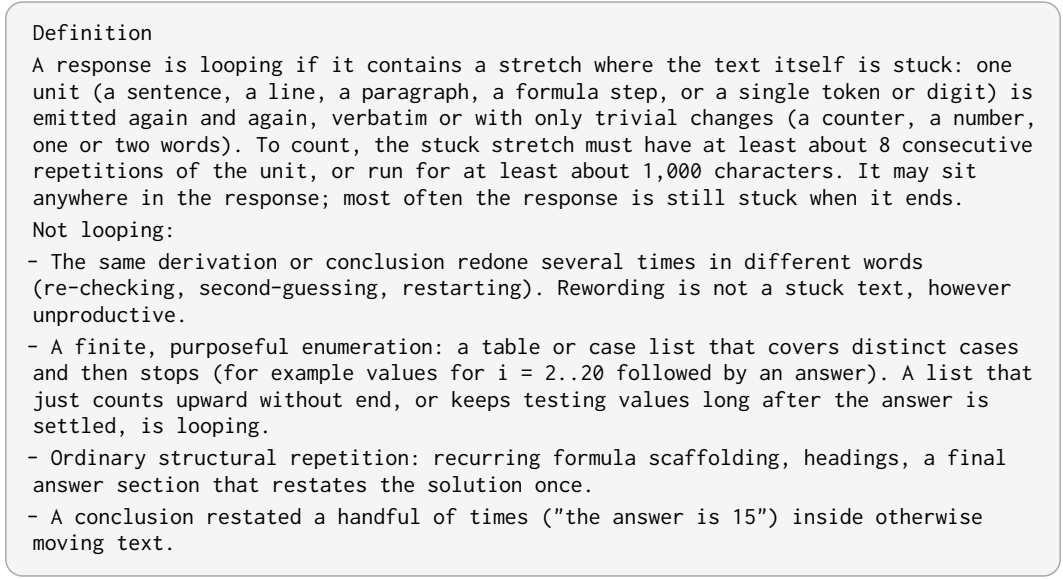

\begin{tcolorbox}[
  enhanced,
  colback=black!4,
  colframe=black!32,
  boxrule=0.6pt,
  arc=6pt,
  left=7pt,right=7pt,top=5pt,bottom=5pt,
  fontupper=\ttfamily\small,
  before upper={\raggedright\setlength{\parindent}{0pt}\setlength{\parskip}{2pt}},
]
Definition

A response is looping if it contains a stretch where the text itself is stuck: one unit (a sentence, a line, a paragraph, a formula step, or a single token or digit) is emitted again and again, verbatim or with only trivial changes (a counter, a number, one or two words). To count, the stuck stretch must have at least about 8 consecutive repetitions of the unit, or run for at least about 1,000 characters. It may sit anywhere in the response; most often the response is still stuck when it ends.

Not looping:\par
- The same derivation or conclusion redone several times in different words (re-checking, second-guessing, restarting). Rewording is not a stuck text, however unproductive.\par
- A finite, purposeful enumeration: a table or case list that covers distinct cases and then stops (for example values for i = 2..20 followed by an answer). A list that just counts upward without end, or keeps testing values long after the answer is settled, is looping.\par
- Ordinary structural repetition: recurring formula scaffolding, headings, a final answer section that restates the solution once.\par
- A conclusion restated a handful of times (\textquotedbl{}the answer is 15\textquotedbl{}) inside otherwise moving text.

\end{tcolorbox}
\caption{Labeling instructions for identifying repetitive errors (excerpt). ``Looping'' in the instructions denotes the repetitive-error category of this paper.}
\label{fig:classification-prompt}
\end{figure}

\hyperref[fig:generation-outcomes]{Figure~\ref*{fig:generation-outcomes}} uses seed-42 responses, with correctness determined by the benchmark scorers. Claude Sonnet 5.5~\citep{anthropic2026sonnet55} with high reasoning effort labels each incorrect response in two independent passes, with model identities withheld; \hyperref[fig:classification-prompt]{Figure~\ref*{fig:classification-prompt}} gives the definition of a repetitive error. The passes agree on 93.6\% of the MATH-500 responses (Cohen's $\kappa=0.87$) and on 95.9\% of the AIME responses ($\kappa=0.88$). Disagreements and doubtful cases go to author review: the authors read them in full and set the final labels.

\section{Truncation Rates of the Main Results}
\label{app:truncation-main}
\hyperref[tab:main-results-truncation]{Table~\ref*{tab:main-results-truncation}} reports the percentage of responses terminated by the output-token limit for the QAT methods in \hyperref[tab:main-results]{Table~\ref*{tab:main-results}}; lower is better, and Avg.\ is the unweighted mean across the nine benchmarks. \ours{} has the lowest average truncation rate in all six model/bit settings.
\begingroup
\def\qatTruncationOnly{}

\endgroup

\section{Truncation Rates across Recovery Stages}
\label{app:truncation-stages}
\hyperref[tab:opd-gain-truncation]{Table~\ref*{tab:opd-gain-truncation}} tracks truncation rates across the recovery stages in \hyperref[tab:opd-gain]{Table~\ref*{tab:opd-gain}}. Calibration lowers average truncation at every model size and bit width, and on-policy distillation further reduces it in all six settings.
\begingroup
\def\qatTruncationOnly{}

\endgroup

\section{Variation across Evaluation Seeds}
\label{app:qat-run-sd}
\hyperref[tab:qat-run-sd]{Table~\ref*{tab:qat-run-sd}} reports sample standard deviations across the three evaluation seeds for each QAT and \ours{} result in \hyperref[tab:main-results]{Table~\ref*{tab:main-results}}, using the divisor $n-1$ with $n=3$. For Avg., we first average the nine benchmark scores within each seed, then compute the SD across seeds.
\begin{table}[!htbp]
\centering
\caption{Sample standard deviations (percentage points) of QAT and \ours{} scores in \hyperref[tab:main-results]{Table~\ref*{tab:main-results}} over evaluation seeds 42, 43, and 44.}
\label{tab:qat-run-sd}
\begingroup
\qatTableSetup
\small
\setlength{\tabcolsep}{0pt}
\begin{tabularx}{\textwidth}{Y}
\toprule
\qatTableHead{Method} & \qatTableHead{W/A} & GSM8K & \makecell{MATH-\\500} & \makecell{AIME\\24--26} & \makecell{Human\\Eval+} & MBPP+ & \makecell{LiveCode\\Bench} & \makecell{MMLU-\\Redux} & \makecell{GPQA-\\Diamond} & IFEval & \qatTableHead{Avg.} \\
\qatTableHeaderRule
\multicolumn{12}{l}{\qatTableModel{Qwen3-1.7B}} \\
EfficientQAT & 3/16 & 0.35 & 1.00 & 1.70 & 2.82 & 1.76 & 0.61 & 0.10 & 4.08 & 4.44 & 0.97 \\
BitDistiller & 3/16 & 0.34 & 1.01 & 2.31 & 1.61 & 0.95 & 0.43 & 0.59 & 2.02 & 1.61 & 0.42 \\
ReasoningQAT & 3/16 & 0.30 & 0.20 & 2.22 & 1.86 & 2.65 & 0.25 & 0.08 & 2.67 & 0.93 & 0.19 \\
\qatTableShade\textbf{OnQ} & 3/16 & 0.65 & 1.42 & 2.31 & 1.53 & 0.55 & 0.63 & 0.26 & 0.77 & 0.21 & 0.49 \\
\midrule
EfficientQAT & 2/16 & 1.59 & 2.75 & 0.64 & 1.83 & 0.85 & 0.81 & 0.49 & 1.82 & 3.49 & 0.80 \\
BitDistiller & 2/16 & 0.50 & 3.48 & 0.00 & 2.54 & 1.19 & 0.36 & 0.96 & 2.31 & 1.11 & 0.56 \\
ReasoningQAT & 2/16 & 0.43 & 1.39 & 0.64 & 2.54 & 1.85 & 0.43 & 0.20 & 2.28 & 0.49 & 0.48 \\
\qatTableShade\textbf{OnQ} & 2/16 & 0.90 & 0.31 & 2.57 & 3.47 & 1.37 & 0.43 & 0.23 & 1.54 & 0.38 & 0.47 \\
\qatTableModelRule
\multicolumn{12}{l}{\qatTableModel{Qwen3-4B}} \\
EfficientQAT & 3/16 & 0.32 & 1.17 & 3.90 & 3.05 & 1.46 & 0.72 & 2.12 & 4.91 & 1.64 & 0.86 \\
BitDistiller & 3/16 & 0.49 & 0.42 & 2.22 & 3.36 & 1.50 & 1.09 & 0.36 & 2.02 & 0.95 & 0.78 \\
ReasoningQAT & 3/16 & 0.37 & 1.14 & 2.94 & 2.79 & 0.70 & 0.20 & 0.12 & 1.82 & 1.26 & 0.43 \\
\qatTableShade\textbf{OnQ} & 3/16 & 0.27 & 0.53 & 1.70 & 2.14 & 0.55 & 0.28 & 0.49 & 4.80 & 1.05 & 1.10 \\
\midrule
EfficientQAT & 2/16 & 0.77 & 0.95 & 2.31 & 2.14 & 1.21 & 1.22 & 0.25 & 5.92 & 3.27 & 1.39 \\
BitDistiller & 2/16 & 0.81 & 0.53 & 2.80 & 1.22 & 1.22 & 1.19 & 0.67 & 2.53 & 1.19 & 0.88 \\
ReasoningQAT & 2/16 & 0.16 & 1.22 & 2.94 & 2.75 & 1.36 & 0.25 & 0.27 & 0.77 & 0.49 & 0.62 \\
\qatTableShade\textbf{OnQ} & 2/16 & 0.54 & 0.31 & 2.31 & 2.31 & 1.50 & 0.28 & 0.09 & 0.77 & 0.83 & 0.43 \\
\qatTableModelRule
\multicolumn{12}{l}{\qatTableModel{Qwen3-8B}} \\
EfficientQAT & 3/16 & 2.47 & 1.04 & 1.70 & 9.37 & 1.25 & 0.49 & 1.38 & 2.49 & 3.55 & 2.17 \\
BitDistiller & 3/16 & 0.08 & 0.81 & 1.70 & 0.35 & 0.40 & 0.72 & 0.16 & 0.00 & 0.49 & 0.30 \\
ReasoningQAT & 3/16 & 0.12 & 0.31 & 3.90 & 1.27 & 1.59 & 1.88 & 0.12 & 2.54 & 0.74 & 0.67 \\
\qatTableShade\textbf{OnQ} & 3/16 & 0.27 & 0.20 & 2.80 & 0.00 & 1.00 & 0.47 & 0.22 & 3.58 & 0.95 & 0.70 \\
\midrule
EfficientQAT & 2/16 & 1.61 & 0.92 & 5.59 & 2.20 & 1.99 & 1.16 & 1.61 & 2.87 & 3.37 & 1.13 \\
BitDistiller & 2/16 & 0.20 & 1.40 & 1.28 & 1.06 & 0.26 & 0.47 & 0.25 & 1.82 & 1.61 & 0.42 \\
ReasoningQAT & 2/16 & 0.45 & 1.15 & 2.57 & 1.96 & 3.23 & 0.77 & 0.66 & 0.87 & 1.12 & 1.09 \\
\qatTableShade\textbf{OnQ} & 2/16 & 0.15 & 0.00 & 4.49 & 1.06 & 1.10 & 0.38 & 0.22 & 2.31 & 0.55 & 0.42 \\
\bottomrule
\end{tabularx}
\endgroup
\end{table}

\end{document}